\documentclass[sn-mathphys]{sn-jnl}

\usepackage{makecell}
\usepackage{siunitx}
\usepackage{algorithm, algorithmicx}
\usepackage{algpseudocode}
\usepackage{graphicx}
\usepackage{anyfontsize}
\usepackage[figuresright]{rotating}
\jyear{2023}%

\theoremstyle{thmstyleone}%
\theoremstyle{thmstyletwo}%

\theoremstyle{thmstylethree}%

\begin{document}

\title[\textcolor{black}{NATMACHINTELL\_A23037798A}]{\textcolor{black}{Bridging the Gap between Chemical Reaction Pretraining and Conditional Molecule Generation with a Unified Model}}


\author[1]{\fnm{Bo} \sur{Qiang}}\email{bqiang@bjmu.edu.com}
\author[1]{\fnm{Yiran} \sur{Zhou}}\email{yrzhou@bjmu.edu.cn}
\author[1]{\fnm{Yuheng} \sur{Ding}}\email{yh\_ding@bjmu.edu.cn}
\author[1]{\fnm{Ningfeng} \sur{Liu}}\email{liuningfengdede@pku.edu.cn}
\author[1]{\fnm{Song} \sur{Song}}\email{ssong@bjmu.edu.cn}
\author[1]{\fnm{Liangren} \sur{Zhang}}\email{liangren@bjmu.edu.cn}
\author*[2]{\fnm{Bo} \sur{Huang}}\email{huangbo@stonewise.cn
}
\author*[1]{\fnm{Zhenming} \sur{Liu}}\email{zmliu@bjmu.edu.cn}

\affil*[1]{\orgdiv{State Key Laboratory of Natural and Biomimetic Drugs, School of Pharmaceutical Sciences}, \orgname{Peking University}, \orgaddress{\city{Beijing}, \postcode{100191}, \country{China}}}
\affil*[2]{\orgname{Beijing StoneWise Technology Co Ltd.}, \orgaddress{\city{Beijing}, \postcode{100080}, \country{China}}}



\abstract{Chemical reactions are the fundamental building blocks of drug design and organic chemistry research. In recent years, there has been a growing need for a large-scale deep-learning framework that can efficiently capture the basic rules of chemical reactions. In this paper, we have proposed a unified framework that addresses both the reaction representation learning and molecule generation tasks, which allows for a more holistic approach. Inspired by the organic chemistry mechanism, we develop a novel pretraining framework that enables us to incorporate inductive biases into the model. Our framework achieves state-of-the-art results on challenging downstream tasks. By possessing chemical knowledge, our generative framework overcome the limitations of current molecule generation models that rely on a small number of reaction templates. In the extensive experiments, our model generates synthesizable drug-like structures of high quality. Overall, our work presents a significant step toward a large-scale deep-learning framework for a variety of reaction-based applications.}

\keywords{Chemical Reaction, Large-scale Pretraining, Molecule Generation}



\maketitle

\section{Main}\label{sec:main}


\textcolor{black}{Deep learning models have found applications across a multitude of scientific research domains~\citep{bert, jumper2021highly, madani2023large}. Pretraining frameworks~\citep{hendrycks2020pretrained, gu2021domain} facilitate the seamless integration of new tasks, thereby expediting the modeling process, especially for scenarios with limited labeled data.}

Chemical reactions are the foundation of drug design and organic chemistry studies. Currently, data-mining works~\citep{lowe2012extraction, lowe2017chemical} have enabled deep learning models to be applied to chemical reactions. 
Based on these data, there have been plenty of data-driven works that intend to delve into the representation learning of chemical reactions. Representation learning refers to automatically learning useful features from the data, which can then be used for various downstream tasks~\citep{goodfellow2016deep}. In earlier works, traditional molecular fingerprints were applied directly for reaction representations\citep{schneider2015development, probst2022reaction}. Inspired by natural language processing (NLP) methods, researchers also applied attention-based network\citep{schwaller2021mapping, irwin2022chemformer} or contrastive learning techniques\citep{wen2022improving, wang2021chemical} in chemical reaction pretraining networks. These representations have been tested on classification tasks\citep{namerxn} or regression tasks\citep{schwaller2021prediction}. However, these methods ignore the fundamental theories in organic chemistry, which limits their performance. For example, electronic effects and inductive effects will be ignored if bonds or atoms outside the reactive centers are masked~\citep{wen2022improving}.

Except for reaction classification tasks, molecule generation based on chemical reactions is also an important application. This branch of models has been proven to be capable of generating synthetically accessible molecules.\citep{korovina2020chembo, button2019automated, gao2021amortized, pmlr-v162-noh22a}. Earlier works always applied a step-wise template-based molecule generation strategy. 
These template-based methods highly depend on predefined building blocks and reactions, which narrow down the accessible chemical space. Similar trends are found in the field of reaction product prediction, in which template-based methods\citep{coley2017prediction} cannot be extrapolated to complex reactions, and this problem is solved by using template-free methods\citep{jin2017predicting, schwaller2019molecular}. In the reaction-based molecule generation task, template-free methods\citep{bradshaw2019model, bradshaw2020barking} have also demonstrated advantages in generalization over templated-based methods. \textcolor{black}{Nevertheless, the template-free molecule generative methods are only capable of generating molecules based on predefined reactants libraries.} \textcolor{black}{In addition to that, it is more favorable to utilize chemical reactions as editing tools to modify the given structure, regarding the hit-to-lead or lead-optimization phase in drug design. The generated chemical library will focus on a subset of chemical space that could be synthesized with fewer reaction steps.}


\textcolor{black}{
In this paper, we present a novel and comprehensive deep-learning framework for chemical reactions. Our framework is designed to address two fundamental tasks: self-supervised representation learning and conditional generative modeling. Unlike existing approaches, we propose a set of meticulously crafted self-supervised tasks specifically tailored for chemical reactions. These tasks include active center prediction, main-reactant sub-reactant pairing, and reactant-product pairing. Through extensive evaluations of challenging reaction tasks, our method surpasses the state-of-the-art, demonstrating its ability to effectively capture domain knowledge in chemical reactions. The promising results obtained pave the way for a wide range of downstream applications.
}


\textcolor{black}{
By efficiently capturing chemical rules, our model is well-suited for generative tasks. Unlike conventional approaches that rely on selecting fragments from predefined reactant libraries, our model takes molecular structures as input conditions and produces representations of corresponding reactants while preserving permutation invariance within reactions. Leveraging the power of a dense vector similarity searching package, our model enables efficient retrieval of reactants from a large reactants/ reagents library. Subsequently, a reaction prediction model is employed to generate product outputs. In comparison to template-based methods that explore only a limited subset of the chemical space, our approach demonstrates superior performance in generating a broader range of synthetically accessible drug-like structures. This characteristic makes our method particularly suitable for virtual library enumeration, as supported by comprehensive statistical analyses and the case study.
}

\section{Results and Disscusion}\label{sec:reults}
\subsection{Challenges in Chemical Reaction Modeling}
 There are several ways to convert chemical reactions into machine-readable structural data. We define the chemical reactions in the following form:
\begin{equation}
    \{ \text{Reactant}_{\text{i}}\}, \text{i}\in m ~{\xrightarrow{\{\text{Reagent}_{\text{j}}\}, \text{j}\in n}} ~\text{Product}
\end{equation}
Chemical reactions involve three main components: reactants, reagents, and products. Reactants are structures that contribute certain substructures to form the products, and the reactant whose atoms are maximally matched with the product is defined as the main reactant. Other reactants are denoted as sub-reactants. Reagents are chemical entities that do not map to any atoms in the product structures but are necessary \textcolor{black}{for providing a certain chemical environment}, such as solvents or acids. To jointly model the probability of reactants, reagents, and products, \textcolor{black}{there are mainly 3 challenges}. 

\textcolor{black}{Firstly, complicated organic chemical mechanisms are hard to model. However, we can sum up these mechanisms with a simpler proposition: If we change the sub-reactants or reagents in an optimized chemical reaction, there is a high likelihood that the reaction will no longer be optimized. This proposition summarized underlining rules within reaction data and enables us to pretrain our model using the contrastive object.}

\textcolor{black}{Secondly, it is imperative for the reactants and reagents to exhibit permutation invariance during the modeling process; however, a significant number of models have disregarded this essential aspect.}
The final challenge is that reagents and reactants play different roles in chemical reactions, which makes the modeling challenging.

\begin{figure}[t]
\centering
\includegraphics[width=1\textwidth]{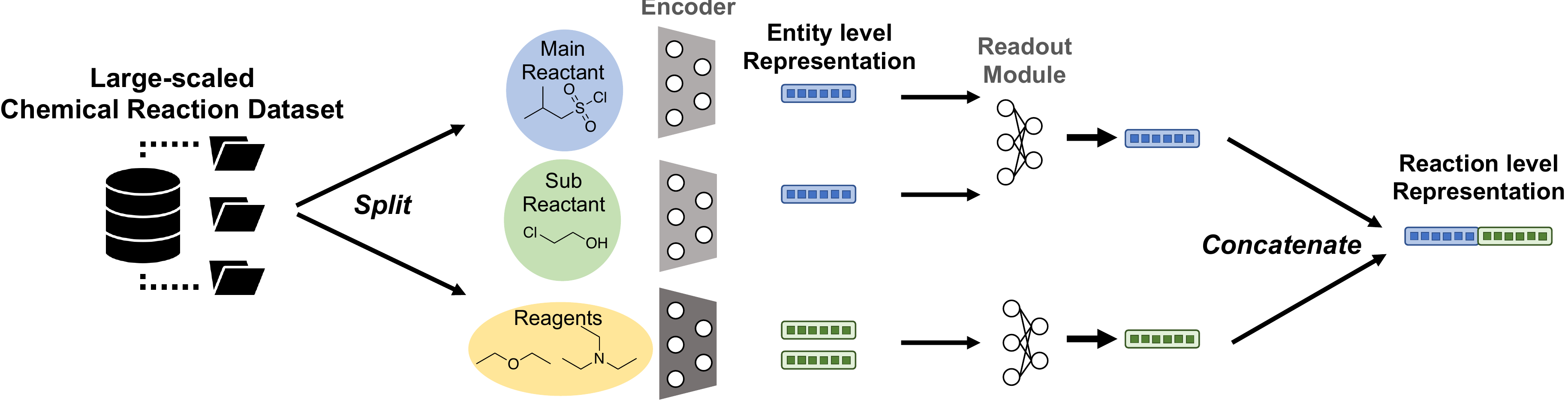}
\caption{An overview of the unified framework of Uni-RXN. The atom mapping is utilized for splitting the reactants into the main reactant and sub-reactants. Then, a multimodal graph/text-based transformer model is applied for deriving the chemical entity-level representations. Entity-level representations are further fused into reaction-level representations. } \label{over} 
\end{figure}

In order to address the above challenges, we design a novel unified framework for modeling chemical reactions. The first challenge comes from the reaction's complicated underlying mechanism is solved by contrastive learning and reactive center prediction task, which will be discussed in detail in Sec.~\ref{contrast}. The second equivariant challenge is solved by shared parameters in encoding as in Fig.~\ref{over} and a permutation invariant generative network in the \textcolor{black}{generative process}. The last challenge is solved by applying a multimodal network for reactants and reagents, which extract information in different manners. Specifically, a graph-based transformer is applied to process reactants and products, and a text-based transformer is applied to process reagents. We will discuss the pretraining and generative models respectively in Sec.~\ref{sec:method}

\subsection{Pretraining Framework and Downstream Tasks}

\begin{figure}[t]
\centering
\includegraphics[width=1\textwidth]{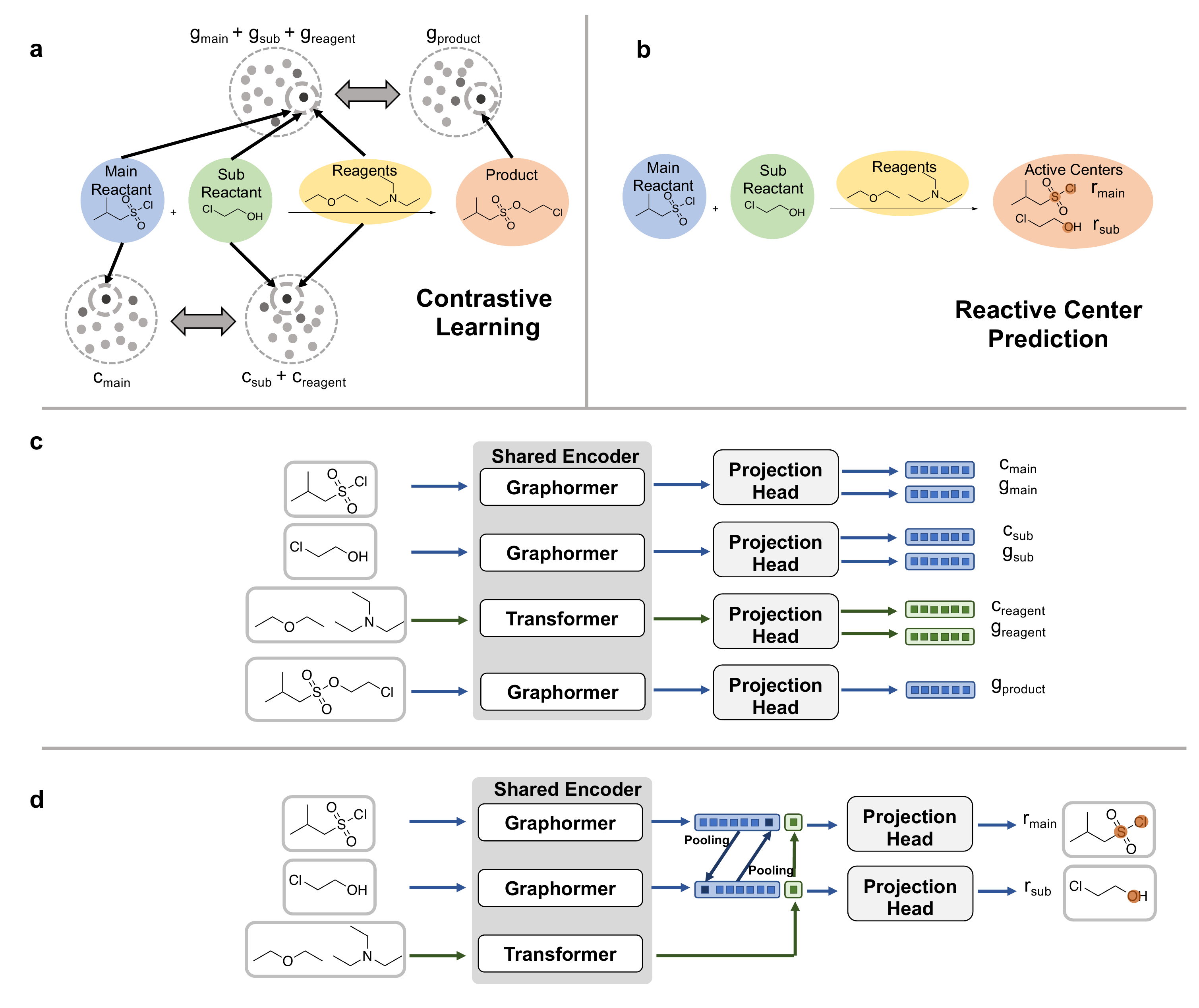}
\caption{\textbf{(a)} Two contrastive learning tasks we utilized for pretraining. The main reactant,  sub-reactants, and reagents are projected into the embedding space using different attention-based model architectures. The similarity between the embeddings of \{main reactant\} and \{sub reactants, reagents\} is maximized because of the underlying organic mechanism. The similarity between the embeddings of \{main reactant, sub reactants, reagents\} and \{product\} is maximized because of the graph-level mapping. \textbf{(b)} An illustration of the reactive center prediction task. \textbf{(c)} The model architecture for contrastive learning. $c$ stands for the chemical training signal which is applied in the first contrastive learning objective. While $g$ stands for the graph training signal which is applied in the second contrastive learning objective.. \textbf{(d)} The model architecture for reactive center prediction tasks. An additional graph-based transformer model is applied to identify the place where chemical bonds are broken or newly formed in chemical reactions.} \label{fig:pretrain_all} 
\end{figure}

\subsubsection{Self-supervised Contrastive Learning}\label{contrast}
 The key components of contrastive learning are the methods of negative data sampling. Rather than dropping or masking atoms outside the reactive center, which would result in a loss of information, we adopt a different strategy illustrated in Fig.~\ref{fig:pretrain_all} \textbf{a}. Our model tends to encode two fundamental aspects of chemical reactions.

Firstly, we model the interactivity between the main reactant and the combination of sub-reactant and reagents (denoted as 2-tuples {sub-reactants, reagents} below). It is well-recognized that the dataset of the chemical reactions is biased. Specifically, only optimized and widely used chemical reactions are curated in the public patent datasets. Therefore, the model only trained on the public dataset will fail to capture any information of negative data (i.e., invalid reactions). To address this, we apply the contrastive loss on reactants and reagents where the negative samples are generated by the random permutation of sub-reactants and reagents among the positive reactions (See Sec.~\ref{sec:method}). We use Information Noise-Contrastive Estimation (infoNCE) as the training objective, projecting the embeddings of the main reactant and \{sub-reactants, reagents\} into the same embedding space with different MLP projection models. When there are multiple sub-reactants or reagents a read-out module is applied. This approach maximizes the similarity between paired main reactants and \{sub-reactants, reagents\} in the embedding space. The training objective is inspired by the powerful generative pretraining model GPT-2/3~\citep{gpt2} where the main reactant and {sub-reactants, reagents} in our model mimic its context and next time-step token, respectively, which enables us to develop the conditional generative model that is discussed in detail in Sec.~\ref{gen_sec}.

Secondly, we seek to model the functional group rearrangement and structure transformation between the combination of main reactant, sub-reactants, and reagents (denoted as 3-tuples {main reactant, sub-reactants, reagents}) and product. Under the chemical condition provided by the reagents, the functional group within the main reactant and sub-reactants are rearranged. In order to learn this transformation process, based on the same encoder, we apply another set of projection heads to predict the embeddings of \{main reactant, sub reactants, reagents\} and product. A similar training process is performed as the first contrastive learning task, in which the paired similarity between \{main
reactant, sub-reactants, reagents\} and product in the embedding space is maximized.

By leveraging these two fundamental aspects of chemical reactions, our contrastive learning framework is able to learn from biased and unoptimized data and generate rich representations.

\subsubsection{Reactive Center Predition}
 Aside from contrastive learning, our model is also trained to predict the reactive centers in chemical reactions as in Fig.~\ref{fig:pretrain_all}\textbf{b}. In our work, we define atoms as chemical reactive centers if they undergo chemical state change. The chemical state is defined as the formal charge, the hybridization, and the neighbor atom types of a certain atom. We propose to use another graph-based transformer model instead of MLP as the projection head, which will be discussed in detail in Section~\ref{sec:method}. This pretraining task further helps our model to understand the position effect in chemical reactions, which is always ignored in related works.

\begin{table}[h]
\begin{center}
\begin{minipage}{\textwidth}
\caption{\textcolor{black}{The accuracy of the chemical reaction classification}}\label{react_class}
\begin{tabular*}{\textwidth}{@{\extracolsep{\fill}}ccccc@{\extracolsep{\fill}}}
\toprule%
 \makecell{Reaction Number \\per Class} & Rxnrep  & MolR  & DRFP  & Uni-RXN \\
\midrule
4 & $0.169\pm0.0227$ & $0.249\pm0.0165$ & $0.258\pm0.0382$ & $\mathbf{0.587\pm0.0229}$ \\
8 & $0.225\pm0.0080$ & $0.328\pm0.0173$ & $0.343\pm0.0159$ & $\mathbf{0.680\pm0.0256}$ \\
16 & $0.305\pm0.0106$ & $0.429\pm0.0213$ & $0.424\pm0.0108$ & $\mathbf{0.754\pm0.0170}$ \\
32 & $0.375\pm0.0159$ & $0.526\pm0.0076$ & $0.498\pm0.0059$ & $\mathbf{0.806\pm0.0101}$ \\
64 & $0.439\pm0.0078$ & $0.615\pm0.0022$ & $0.559\pm0.0044$ & $\mathbf{0.841\pm0.0050}$ \\
128 & $0.489\pm0.0036$ & $0.692\pm0.0025$ & $0.617\pm0.0059$ & $\mathbf{0.865\pm0.0030}$ \\
\botrule
\end{tabular*}
\footnotetext{The accuracy is computed multiple times on different random samples. The Standard Deviation(std) of the accuracy is listed after $\pm$. The higher accuracy indicates better performance. \textcolor{black}{Trained graph neural network encoder from Rxnrep~\citep{wen2022improving}, MolR~\citep{wen2022improving} is applied to compute the baseline model representation. The package that computes DRFP~\citep{probst2022reaction} representation is downloaded directly from rxn4chemistry}}
\end{minipage}
\end{center}
\end{table}

\subsubsection{Reaction Classification}
 After completing the two pretraining tasks on the USPTO MIT dataset~\citep{jin2017predicting}, we use the shared encoder to generate representations for downstream tasks. To ensure a fair comparison, we do not split the main reactants and sub-reactants when generating representations. Most previous works evaluated their representation on TPL 1k dataset~\citep{schwaller2021mapping} for reaction classification and reach over 90 percent accuracy. Our model provides comparable performance on this dataset, which the results are provided in the SI Tab.~\ref{tpl}. We also chose a more challenging reaction classification dataset, Schneider~\citep{schneider2015development}, \textcolor{black}{which used a comprehensive ontology system to classify reactions into superclass and secondary class. When evaluating our model, we removed the reactions whose templates can be found in our test set from the pretraining stage.} 
When generating the representation, we mask the product of the chemical reactions to prevent the model from relying on simple graph pattern recognition. To make the classification even harder, we balanced the dataset by randomly drawing the same number of reactions for each class both in the training and testing dataset, following previous work~\citep{wen2022improving}. The results on this balanced benchmark dataset are demonstrated in Tab~\ref{react_class}. 

Logistic Regression (LR) Classification offered by the cuML package is applied here as the reaction classification heads based on the chemical reaction representations. We denote our representation as Uni-RXN and compared it with three baseline models. The accuracy of predicting the correct reaction class drastically decreases when the number of reactions per class drops from 128 to 4. In the range of dataset sizes we tested, our model has outperformed the baseline models by a large margin, especially on small training sets. Our model predicts more than half of the reaction classes with only 4 reactions per class for training. Notably, we keep the encoder parameter fixed when comparing our model with other baselines. Our model demonstrates impressive results without finetuning any pretrained parameters. In conclusion, Uni-RXN is a powerful tool to classify chemical reactions even without product information which shows great potential in reaction forward prediction applications.

\bigskip

\begin{figure}[t]
\centering
\includegraphics[width=1.0\textwidth]{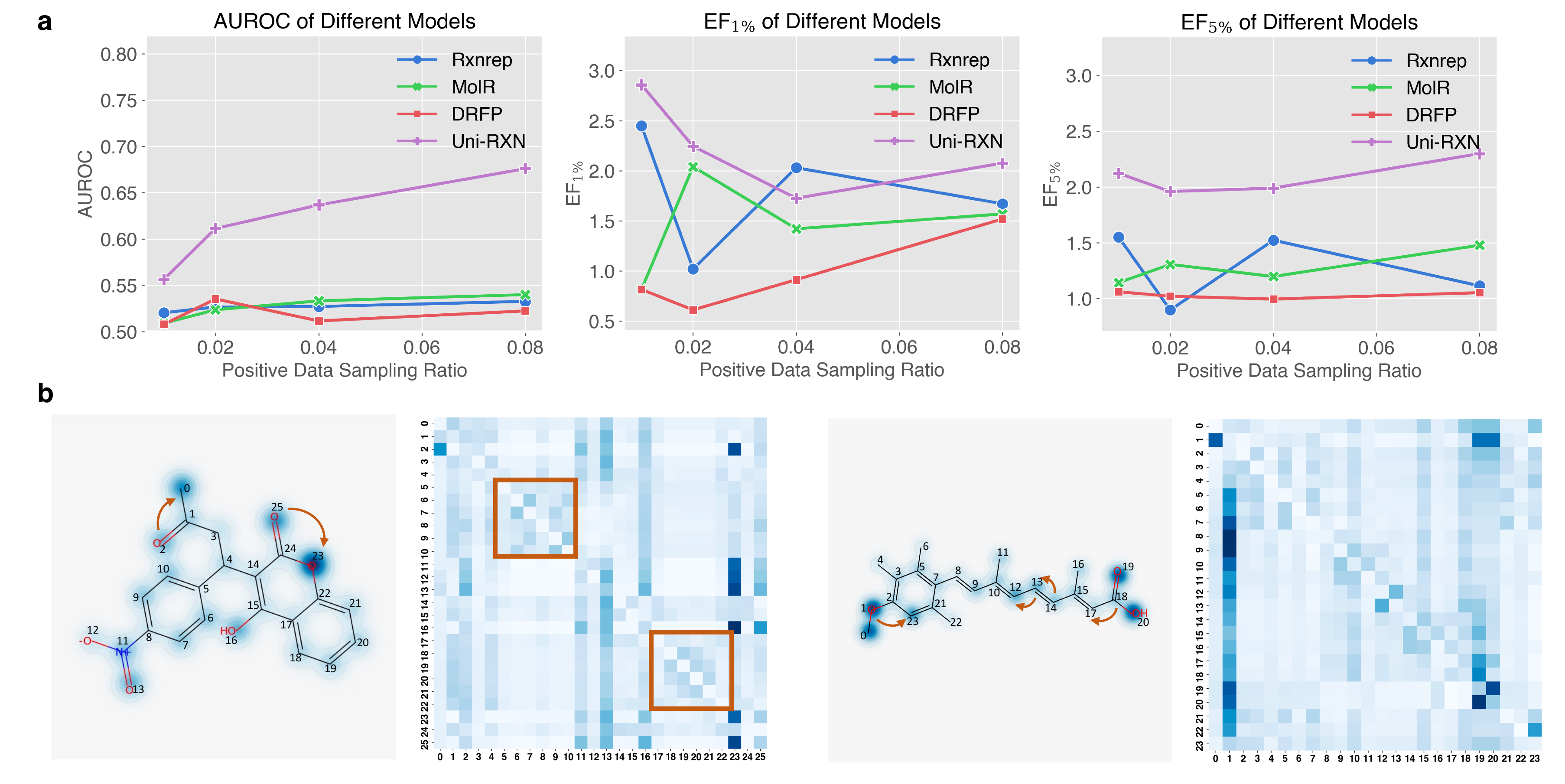}
\caption{\textcolor{black}{\textbf{(a)}The line charts illustrate the AUROC, EF$_1$$_{\%}$ and EF$_1$$_{\%}$ of different representations in the chemical reaction retrieval task. \textbf{(b)}The attention map of the graph-based transformer encoder. Atoms that belong to the same functional group have high cross-attention which demonstrates that our model is capable of learning position effect and identifying reactive atoms.}} \label{pretrain_results}  
\end{figure}

\bigskip

\subsubsection{Reaction Retrieval}
\textcolor{black}{We evaluate our model's ability to distinguish optimized reactions from unoptimized reactions using the reaction retrieval task. This evaluation ensures the effective implementation of our chemical informative representation in the generative model, preventing the generation of suboptimal reactions. We curated a dataset, as described in Section~\ref{sec:method}, comprising positive examples of successful reactions and negative examples of noisy or suboptimal reactions. This dataset serves as a reliable benchmark for assessing our model's capability to identify optimized reactions.}


\textcolor{black}{
We conduct experiments using different positive data sampling ratios ranging from 0.01 to 0.08. These low sampling ratios were chosen to simulate real-world scenarios where only a relatively small proportion of reactions are optimized.  As depicted in Fig.~\ref{pretrain_results}\textbf{a}, our model outperforms other baseline models in most settings. The EF results prove that even if less than 1\% of data are positive results, Uni-RXN is capable of differentiating the optimized reactions from unoptimized ones. We also provide results on a different template-based mechanism~\citep{filter} negative reaction sampling method in the SI Sec.~\ref{si_neg}, where Uni-RXN also outperforms the baseline models. In conclusion, our representation can be applied to reaction retrieval tasks, even with extremely limited positive data, and holds the potential to assist chemists in identifying the correct reagents and sub-reactants for high-yield organic reactions.
}

\subsubsection{Visualization of Attention}
 \textcolor{black}{The visualization of the Transformer attention sheds light on the model's processing of the input graph, providing insights into its performance on the above two important downstream tasks.} To examine this, we present the attention weights in Fig.~\ref{pretrain_results}\textbf{b}. \textcolor{black}{The attention maps clearly illustrate that our model learns to focus on the reactive portion of the input main reactant molecules. }
\textcolor{black}{Notably, heteroatoms in active groups such as the ester group which play crucial roles in chemical reactions, exhibit higher attention scores compared to carbon chains and other heteroatoms. Moreover, our model captures more intricate rules involving interactions between neighboring functional groups. For instance, when a benzene ring possesses a strong electron-withdrawing group, the ortho-position becomes more reactive, and our encoder effectively captures this fundamental chemical rule by directing attention to the ortho-position. Furthermore, the attention maps exhibit well-clustered patterns, with self-attentions between atoms within the same aromatic ring due to their shared electron cloud. Conversely, the attention between neighboring functional groups receives higher scores, as their relationships define the behavior of reactants in organic chemical reactions."}

\subsection{Conditional Generation Framework and Drug-like library design}~\label{gen_sec}
 \textcolor{black}{Our model not only excels in classification tasks but also offers a valuable tool for studying Structure-Activity Relationship (SAR) in medicinal chemistry research. By leveraging the power of our pretrained encoder, we enable the generation of multiple synthesizable analogues from a given hit structure. This streamlined approach provides researchers with a simplified method for exploring SAR and designing focused chemical libraries.}
However, generating analogues through chemical reactions on a seed structure poses a challenge. 
\textcolor{black}{Template-based methods simplify conditional molecule generation by confining sampling in an infinite space to a predefined subspace, reducing the search space. However, limitations arise when the available subspace becomes limited or empty, restricting direct template application.}

\begin{figure}[t]
\centering
\includegraphics[width=1\textwidth]{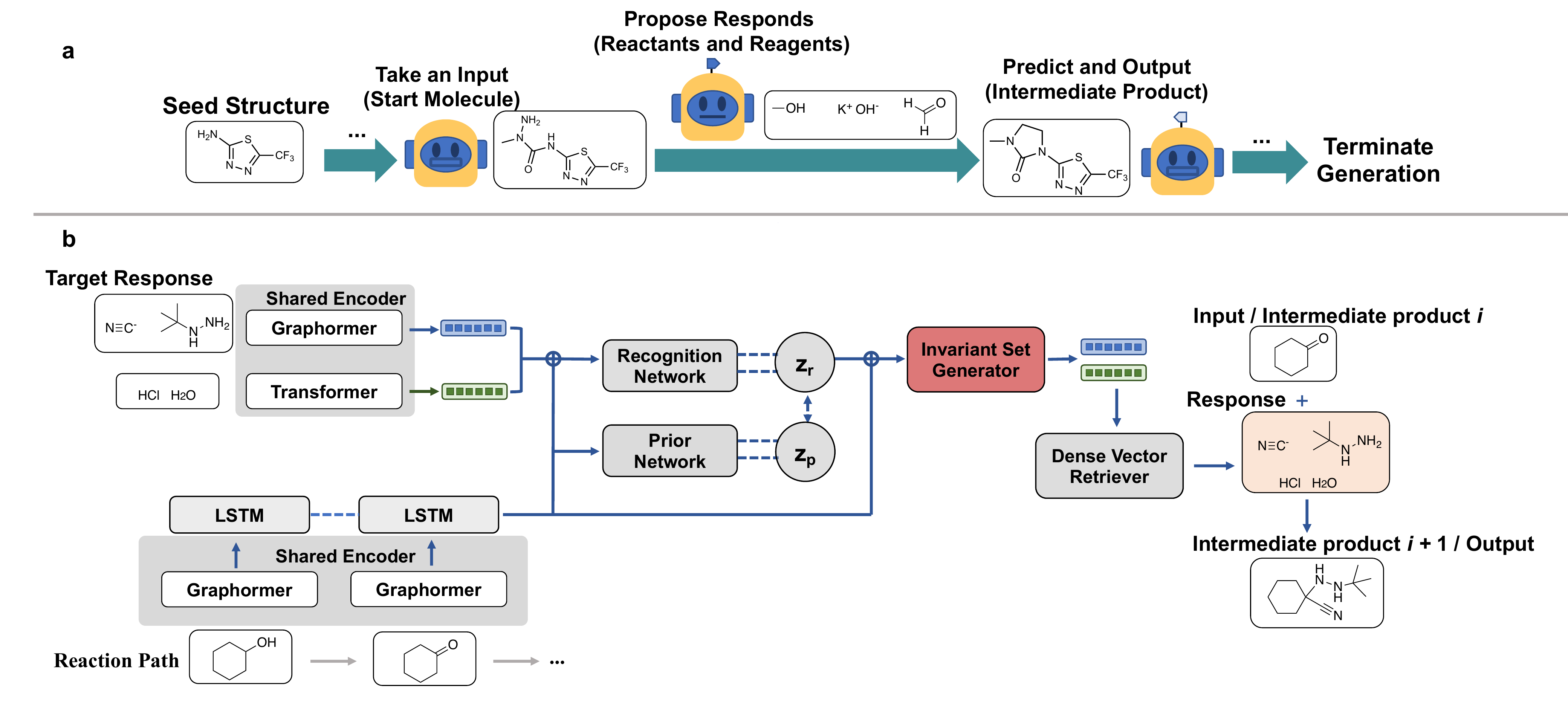}
\caption{\textcolor{black}{\textbf{(a)} An overview of the generation framework. A sequential process is proposed to generate the analogues. At each step, the model proposed the sub-reactants and reagents, then the reaction predictor outputs the product. The product is fed to the model as the input for the next step until the termination criterion is met.\textbf{(b)} The model architecture of the generative model. The pre-trained model is utilized here as the encoder for target structures and main reactant inputs.}} \label{gen_all} 
\end{figure}

\textcolor{black}{To overcome this challenge, we develop a template-free generative model that efficiently generates chemical reaction paths. Each path consists of a series of reactions where the product of the previous reaction is the main reactant of the subsequent reaction. Such paths simulate human experts who utilize chemical reactions to expand the chemical space based on the seed molecule. A conditional variational encoder network, denoted as Uni-RXN$_\text{Gen}$ is trained to generate reaction paths autoregressively by approximating the likelihood of sub-reactants and reagents based on the reaction path from previous steps, as illustrated in Fig.~\ref{gen_all}\textbf{a}.}

\textcolor{black}{The architecture of our model is depicted in Fig.~\ref{gen_all}\textbf{b}. Instead of generating the sub-reactants and reagents directly, we generate the representations of these molecules' structures. Two separate encoders extract the information from the reaction path condition and the target responses. Then the invariant generator decodes the latent variable to generate the target representations. After Uni-RXN$_\text{Gen}$ generates the target representations, a dense vector retriever is used to search for reactants and reagents in a large commercially accessible molecule library. Based on the input main reactant and the retrieved sub-reactants and reagents, another network predicts the product of the proposed new reactions. In short, Our model provides an efficient and effective workflow for generating chemical analogues by sampling reactions and predicting the results sequentially.} 

To evaluate our model's capacity of generating similar molecule structures conditioned on the input seed molecules, we use \textcolor{black}{2567} structures from the Drugbank database~\citep{wishart2018drugbank} to derive large drug-like datasets using our generative model. \textcolor{black}{We compared our model with 4 baseline models, SynNet~\citep{gao2021amortized}, Lib-INVENT~\citep{libinvent}, DINGOS (\textit{De novo}) and DINGOS (condition)~\citep{button2019automated}.}

\bigskip

\bigskip

\subsubsection{Properties Evaluation}\label{sec:prop}
\begin{figure}[t]
\centering
\includegraphics[width=0.9\textwidth]{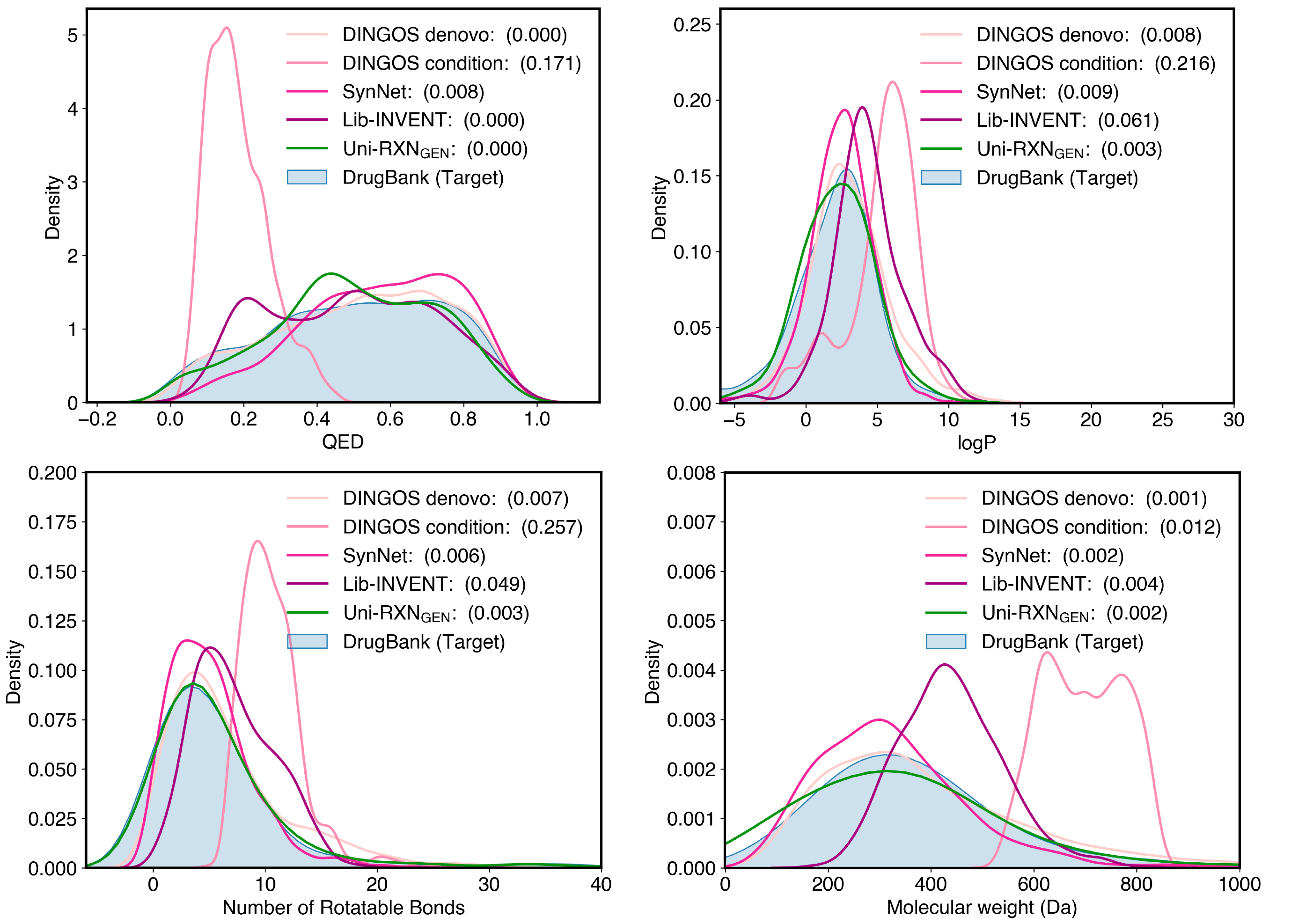}
\caption{\textcolor{black}{The properties of the drug-like molecule generated by different reaction-based molecular generation models. The MMD distribution distances are listed within the ().} } \label{gen_results} 
\end{figure}

 \textcolor{black}{To evaluate the quality of the generated structures, we computed several basic drug-like properties and compared them with real drugs, as illustrated in Fig.~\ref{gen_results}\textbf{a}. Our objective is to generate molecules that closely resemble real drugs in terms of their property data distribution. It is evident that DINGOS(condition) generates molecules with a shifted chemical property distribution despite performing only a few steps of reaction modification on the seed molecules. Regarding the metrics of Molecular Weight and QED, Uni-RXN$_\text{Gen}$, SynNet, and DINGOS (\textit{De novo}) provide comparable results. However, the baseline methods generate molecules with more lipophilic structures and an increased number of rotatable bonds, unlike our model.}

\begin{figure}[t]
\centering
\includegraphics[width=1.0\textwidth]{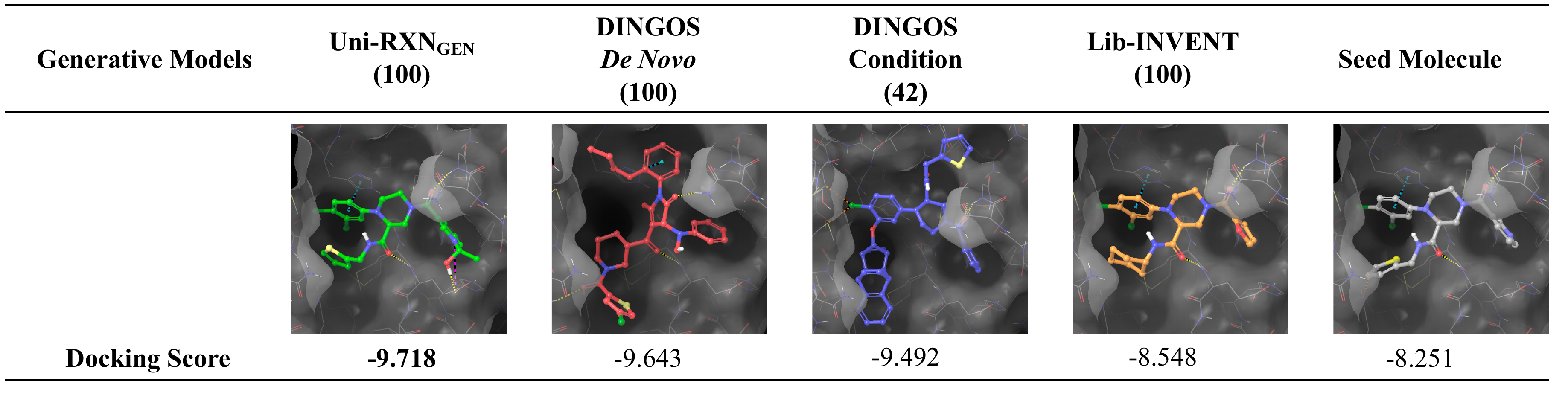}
\caption{\textcolor{black}{The docking pose of the top scoring generated COVID-19 3CLPro inhibitors. the gray poses are the reference poses of the seed molecule derived from the original PDB file. The scaffold of the Uni-RXN sampled molecule and Lib-invent generated molecule aligned perfectly with the reference pose and the Uni-RXN sampled molecule has a higher absolute docking score.}} \label{fig:case_study} 
\end{figure}

This paper goes beyond assessing the drug-likeness of molecules generated by Uni-RXN$_\text{Gen}$, as we also use synthetic accessibility scores to evaluate their synthesizability. To accomplish this, we apply two different metrics (SAScore~\citep{sas}, and RA~\citep{ra}), and the results are presented in Tab.~\ref{sas}.
\textcolor{black}{Among all the methods, our model, DINGOS(condition), and Lib-INVENT are able to generate molecules directly from the input structure, other methods need more reaction steps to generate from scratch.} Unlike DINGOS(condition), which generates larger molecules due to the lack of decomposition reaction in the predefined templates, our generated molecules scored favorably on the SAScore and RA metrics, indicating that they are easier to synthesize than the input seed molecules. Template-based \textit{de novo} methods generate molecules distanced to the seed molecules and have obvious distribution shifts on SAScore and RA. In combination with the distance metrics, it could be observed that DINGOS (\textit{De novo}) and SynNet sacrifice drug similarity and validity for lower synthetic accessibility scores, respectively. It indicates that the template-based methods tend to generate excessively simple molecules, which is undesirable in drug discovery research. Furthermore, our proposed modifications require fewer reaction steps to perform than to carry out a from-scratch complicated route by multiple template reactions. Overall, our approach provides a more effective way to generate molecules based on available drugs, where (1) fewer steps of chemical reactions are required and (2) the generated molecules exhibit a decent balance between drug similarity and synthetic accessibility.

\bigskip
\begin{table}[h]
\begin{center}
\begin{minipage}{\textwidth}
\caption{\textcolor{black}{Evaluation of Synthetic Accessibility Scores, Validity, Chemical Distance and Scaffold Diversity}}\label{sas}
\resizebox{\textwidth}{!}{
\begin{tabular}{@{\extracolsep{\fill}}cccccccc@{\extracolsep{\fill}}}
\toprule%
 ~ & SAScore $\downarrow$   & RA $\uparrow$  & Valid(\%) & \makecell{Chemical\\Distance} & \makecell{Mol\\Diversity} & \makecell{Scaffold\\Entropy}\\
 \midrule
 Seed Molecules & 3.631 &  0.703 & -- & -- & 0.906 & 6.23\\
 SynNet & 2.941 &  0.847 & 66.61 & 0.450 & 0.897 & 5.77 \\
 DINGOS (\textit{De novo}) & 3.080 &  0.788 & 100.00 & 0.602 & 0.899 & 5.53\\
 DINGOS (condition) & 3.845 & 0.636 &  45.81 & 0.824  & 0.424 & 5.60\\
 Lib-INVENT (condition) & 3.710 & 0.558 & 100.00 & 0.681 & 0.869 & 5.37\\
 \midrule
Uni-RXN & 3.643 &  0.725 & 100.00 & 0.362 & 0.925 & 6.09 \\
\botrule
\end{tabular}}
\footnotetext{$\downarrow$ indicates that the lower score means the molecules are easier to synthesize, while $\uparrow$ indicates that the higher score means the molecules are easier to synthesize. \textcolor{black}{Valid indicates the proportion of drug structures the model is capable of generating valid analogues. Chemical Distance measures the ECFP4 distance between the generated molecule and the input. Mol Diversity and Scaffold Entropy measure the diversity on full structure level and scaffold level.}}
\end{minipage}
\end{center}
\end{table}

\textcolor{black}{Validity is an important metric in assessing the performance of a model, as an ideal model should be capable of generating analogues and modifying valid structures based on the seed molecules. However, both SynNet and DINGOS (condition) demonstrate limitations in generating analogues for a limited proportion of input molecules, even though they can generate synthesizable molecules. In addition to validity, we further evaluate molecule diversity and scaffold entropy to assess the models. As a result, there are fewer dominant scaffolds and similar molecules within the molecules generated by Uni-RXN$_\text{Gen}$.}


\subsubsection{SARS-CoV-2 Main Protease Inhibitor Design Case Study}
 Instead of \textit{de novo} designing inhibitors~\citep{morris2021discovery}, we intend to optimize existing ligand structures using our structure conditional generative model. When generating analogues of drug-like molecules, maintaining a stable 3D binding conformation is crucial for ensuring that the newly generated molecules can bind to the same protein pockets.

\textcolor{black}{
To demonstrate that Uni-RXN$_\text{Gen}$ generates molecules that fit into the target protein pocket, we conducted a case study on 3CL$_\text{Pro}$ inhibitor design. In our experiment, we generate analogues based on the seed molecule derive from the original complex~(PDB id: \href{8ACL}{http://doi.org/10.2210/pdb8ACL/pdb}) with our method and other reaction-based generative model~\citep{button2019automated, gao2021amortized} and a library design model, Lib-INVENT~\citep{libinvent} as shown in Tab.~\ref{tab:dock} and Fig~\ref{fig:case_study}(c). Our method outperforms other methods on average docking scores and top docking scores when the same number of molecules are kept. Besides, Uni-RXN$_\text{Gen}$ and Lib-INVENT generate top-scored analogues with similar binding conformations and similar topologies, as suggested by the docking results and the fingerprint distances. However, our model still generates molecules of high diversity which outperforms DINGOS(condition) and Lib-INVENT, which proves that our model is able to explore the chemical space adjacent to the input seed molecule effectively. We also found out that template-based method, \textit{e.g.} DINGOS (condition) is only able to generate 42 valid molecules with the template reactions which proves that reaction templates harm the ML model's ability to explore constrained chemical space.}

These findings demonstrate that our model can aid medicinal chemists in discovering SAR in a more efficient manner by providing a large number of analogues with higher binding affinity.



\begin{table}[h]
\begin{center}
\begin{minipage}{\textwidth}
\caption{\textcolor{black}{The similarity between seed molecules and generated molecules. The docking scores and diversity of the generated analogues}}\label{tab:dock}
\begin{tabular*}{\textwidth}{@{\extracolsep{\fill}}ccccc@{\extracolsep{\fill}}}
\toprule%
&   \makecell{DINGOS \\\textit{De novo}\\(100)} & \makecell{DINGOS \\condition\\(42)} & \makecell{Lib-INVENT\\(100)} & \makecell{Uni-RXN$_\text{Gen}$ \\(100)} \\
\midrule
\makecell{Similarity$_{\text{mean}}$} &  0.2977 & 0.6548 & 0.6942 & 0.7748 \\
\makecell{Similarity$_{\text{top}}$} &  0.3816 & 0.7155 & 0.8163 & 0.8854 \\
\makecell{Diversity} &  0.7913 & 0.3184 & 0.3455 & 0.3949 \\
\makecell{Docking Score$_{\text{mean}}$} & -6.5068 & -7.7953 & -7.5420 & -7.8301 \\

\botrule
\end{tabular*}

\footnotetext{The number of the generated molecules we docked with the protein pocket is listed in the (). We kept the 100 analogues that have the closest Jaccard distance to the input molecules for docking. However, the DINGOS (condition) model is only able to generate 42 valid molecules.}
\end{minipage}
\end{center}
\end{table}

\bigskip

\bigskip

\section{Conclusion}
In this paper, we have presented a novel approach to bridging the gap between reaction-based molecule pretraining and generation tasks. \textcolor{black}{Our approach offers several advantages, including the ability to derive rich representations for challenging chemical reaction classification tasks.} Uni-RXN outperforms other baseline models by a significant margin and achieved 58.7\% accuracy with only 4 data points per class provided. The transformer model can be also applied to \textcolor{black}{differentiate optimized reactions and unoptimized reactions} in chemical reaction data. \textcolor{black}{Additionally, the encoder can be effortlessly applied to structure conditional generation. The experimental results highlight the favorable properties of the molecules generated by our model, making them well-suited for drug discovery tasks.} Our model is capable of generating molecules with more drug-like properties and synthesizable accessibility. Combined with virtual screening methods, such as molecule docking, this generative model enables efficient SAR studies. The vast synthesizable drug-like chemical space that our model generates can improve the true positive rate in drug repurposing or hit molecule searching.

\section{Methods}\label{sec:method}
\subsection{Model Architecture}
\subsubsection{Pretraining Model}
 \paragraph{Transformer Model}\textcolor{black}{Our pretraining network utilizes two attention-based models as encoders to handle the multimodal nature of chemical reaction data.} The reactants and products are encoded by a graph-based transformer model, \textcolor{black}{while} the reagents are translated to sequence representation SELFIES~\citep{krenn2020self} and then encoded by a text-based transformer model. The projection heads in contrastive learning tasks are simple MLPs with 2-layers of hidden parameters to train. \textcolor{black}{We provide a detailed discussion of the two types of transformer models we have designed in the following paragraph.}
 
\textcolor{black}{For the text-based transformer, we base it on the vanilla transformer~\citep{vaswani2017attention}. However, directly applying the transformer to reagents poses challenges because the tokens within chemical entities follow a sequential order, while the entities themselves are order-agnostic. To address this, }we develop a hierarchical transformer to model reagents. In the first step, the sequential tokens are encoded by a vanilla transformer, and the embeddings of the CLS tokens are utilized to represent the entities respectively. Then, a transformer without positional encoding is applied as the readout module from the CLS embeddings of multiple reagents.

The graph-based transformer is inspired by previous work~\citep{ying2021transformers}. Since molecule graphs are permutation invariant, we discard node positional encoding and apply edge encodings to embed the topological structure of the molecules. \textcolor{black}{The edge feature inputs consist of the bond type and an indicator token that signifies whether the bond belongs to a ring. Similarly, for the node feature, we incorporate the Atom element type, the number of formal charges, the hybridization type, and a ring indicator.} The edge features are applied in our model to get the attention map as follows:
\begin{align}
    A_{i j}=\frac{\left(h_{i} W_{Q}\right)\left(h_{j} W_{K}\right)^{T}}{\sqrt{d}}+b_{\phi}\left(e_{ij}\right)+c_{\phi} SP(i j)
\end{align}
The $h_{i}$ and $h_{j}$ denote the node embeddings of node $i$ and node $j$. $d$ is the dimension of the edge features. $e_{ij}$ is the one-hot edge feature (all zeros indicate that there are no edges between them), and $SP$ is a function that returns the shortest path from node $i$ to node $j$. $b_{\phi}$ and $c_{\phi}$ are all learnable neural networks. With this modification in attention mechanism, the model is able to capture and encode local structure while maintaining long-range interactions which are always hard to encode with \textcolor{black}{traditional message-passing neural networks~\citep{zhang2020dynamic}}. This function enables our model to learn the interaction between functional groups and understand more complicated organic chemistry. The model can take multiple graphs as input, so we set the shortest path distance between atoms that do not belong to the same reactant molecules as infinity. In this way, the model automatically aggregates information within this hypergraph, where both local and global interactions are enabled.

When this graph-based transformer model is applied, we add an auxiliary node, the virtual node. It is connected to all atoms even if they belong to different molecules, the embeddings of the virtual token can be used as the results of a full graph-level readout. In the reactive center prediction task, we apply sum pooling to the \textcolor{black}{token level representations from the graph-based transformer} and appended the CLS token from the text-based transformer as the additional input to the prediction model, which is illustrated in Fig.~\ref{fig:pretrain_all}\textbf{d}. This design maintains permutation invariance and allows efficient information exchange between models.

\paragraph{Training Objective}
\textcolor{black}{In this section, we discuss the training objective of pretraining.} In the first contrastive learning task, we want to model the interaction between reactants. Hence, the InfoNCE Loss is employed here to maximize the similarity between the representation of main reactants $x=c_{\text{main}}$ and their chemical reaction environment(the sub-reactants and reagents) $v=c_{\text{sub}}+c_{\text{reagent}}$. If we use $X$ to represent all possible reactants and $V$ to represent all possible chemical reaction environments.
\begin{align}
    \mathcal{L}_{I n f o N C E}=-\mathbb{E}_X\left[\log \frac{f\left(x_{i}, v_{i}\right)}{\sum_{x_{j} \in X} f\left(x_{j}, v_{i}\right)}\right]-\mathbb{E}_V\left[\log \frac{f\left(x_{i}, v_{i}\right)}{\sum_{v_{j} \in V} f\left(x_{i}, v_{j}\right)}\right]
\end{align}
We apply the inner product here as the similarity function $f$. In the second contrastive learning task, the transformation from reactants to products is modeled by our probabilistic model. The same $\mathcal{L}_{I n f o N C E}$ function is applied when $x=g_{\text{main}} + g_{\text{sub}}+g_{\text{reagent}}$ stands for chemical reaction inputs(reactants, reagents) and $v=g_{\text{product}}$ stands for chemical reaction outputs(products). To estimate the loss efficiently, the $X$ and $V$ are approximated by in-batch sampled embeddings. \textcolor{black}{For training the reactive center prediction task, we employ the binary cross entropy loss as our objective function. We apply this to predict multiple reactive centers defined by atoms whose chemical environments have changed.}

\subsubsection{Generative Model}\label{sec:method-gen}
 \paragraph{Conditional Variational Auto-encoder}
We build a conditional variational auto-encoder(CVAE) based on the pre-trained encoder to generate the representation of the responses (sub-reactants and reagents). As illustrated in Fig.~\ref{gen_all}\textbf{b}, we first use a long short-term memory network(LSTM) to collect the embeddings along the reaction paths we sample from the chemical reaction network~\citep{jacob2018statistics}. Two MLP networks are designed for variational inferencing. \textcolor{black}{The recognition network takes both the target representation and the embeddings of the previous reaction path as inputs, while the prior network processes only the latter. By minimizing the Kullback-Leibler (KL) divergence between these dual latent variables, akin to classic variational auto-encoder models, we obtain the latent variable $z$. By minimizing the Kullback-Leibler (KL) divergence between these dual latent variables, akin to classic variational auto-encoder models, we approximate the recognition distribution by our prior network and get latent variable $z_\text{prior}$. Subsequently, the Invariant Set Generator utilizes this latent variable to generate the target representations.} During training, the prior network is trained to approximate the distribution generated from the recognition network. \textcolor{black}{During the generation phase}, only the latent variable from the prior network is passed forward to the generator.

\paragraph{Invariant Set Generator}
\textcolor{black}{
Generating a permutation invariant set presents a challenge, as discussed in prior research~\citep{vignac2021top}. In this study, we propose a similarity-based approach to generate varying numbers of reactant/reagent representations from the latent variable. To accomplish this, we maintain a reference parameter set and employ its angle to determine the selection of data points for representation generation. Let $\boldsymbol{\Theta}$ and $\boldsymbol{R}$ denote the angle and latent variable of the reference set, respectively. The formulation of the model is as follows:}

\begin{align}
\boldsymbol{a} & =\operatorname{MLP}_{1}(\boldsymbol{z}) \nonumber\\
\boldsymbol{c} & =\boldsymbol{\Theta} \boldsymbol{a} / \operatorname{vec}\left(\left(\left\|\boldsymbol{\theta}_{i}\right\|_{2}\right)\right) \nonumber\\
s &= \operatorname{argsort}_{\downarrow}(c[:n_{max}])\nonumber\\
\tilde{\boldsymbol{c}} & =\operatorname{softmax}(\boldsymbol{c}[s]) \nonumber\\
\boldsymbol{X} & =\boldsymbol{R}[\boldsymbol{s}] \odot \tilde{\boldsymbol{c}} W_{1}+\tilde{\boldsymbol{c}} W_{2} \nonumber\\
\boldsymbol{X} &= \operatorname{MLP}_{2}(\boldsymbol{X}, \boldsymbol{z})\nonumber\\
\boldsymbol{Y} &=  \boldsymbol{X}[i_1,\cdots, i_n],~s.t.~ \forall i_x, \operatorname{MLP}_{3}(\boldsymbol{X}[i_{x}] > \delta)\nonumber
\end{align}

The $W_1$, $W_2$ are trainable weight matrices, and $\operatorname{MLP}_{1}$, $\operatorname{MLP}_{2}$, $\operatorname{MLP}_{3}$ are simple trainable MLP networks. $n_{max}$ is the maximum number of responses set in the training set. $\theta$ is the threshold that we apply to decide whether to select the data points. If we use $m$ to represent the dimension of target representations, the output $\boldsymbol{Y}$ will be the $n \times m$ matrix that we need. Instead of using First-N or Top-N set generation methods, we apply a scoring network $\operatorname{MLP}_{3}$ to determine the size of the set we generate, which helps us to generate more diverse results and update more latent variables during training.

\subsubsection{Implementation of Forward Reaction Predictor}
\textcolor{black}{We applied the LocalTransformer~\citep{chen2022generalized} as our predictor, since it is now the state-of-the-art model for forward reaction prediction. However, any forward reaction prediction model can be plugged into our method without much effort.}

\subsection{Experiment Setups}
\subsubsection{Negative Data Construction}~\label{our_neg_method}
 To construct negative samples, we develop a workflow to sample unsuccessful chemical reactions from classified Schneider dataset~\citep{schneider2015development}. \textcolor{black}{Recognizing that different classes of chemical reactions employ distinct sub-reactants and reagents, we employ various techniques to generate negative samples, namely adding, deleting, and switching operations.} 
 
\textcolor{black}{In the adding operation, we randomly sample sub-reactants and reagents from reactions belonging to other reaction classes and append them to the reactions to be perturbed. } Conversely, in the deleting operation, random numbers of sub-reactants and reagents are removed. \textcolor{black}{As for the switching operation, it involves replacing random numbers of sub-reactants and reagents with chemical entities from other reaction classes. Subsequently, we apply a filtering process to refine these perturbed reactions.} If the forward reaction product predictor succeeds in generating \textcolor{black}{exactly the same valid product} for our perturbed reactions, \textcolor{black}{they are excluded from the negative sample dataset.} At last, all remaining perturbed reactions are kept and \textcolor{black}{reatain as} the negative data points. 


\textcolor{black}{In addition to our approach, we have also adopted the 'random template' negative sampling strategy from a previous work\citep{filter} where reaction templates are applied to select sub-reactants. }The results of this strategy are presented in the Supplement Information~\ref{si_neg}.

\subsubsection{Property Evaluation}
 We examine some important drug-like properties and synthetic accessibility scores to prove that our generative model is able to derive drug-like molecules based on drug structures. \textcolor{black}{
Note that for DINGOS, Lib-INVENT and Uni-RXN$_\text{Gen}$, multiple structures are generated for the same input. To ensure quantity consistency, we retained the most similar structure within the first 8 sampled structures, as measured by ECFP4.}

Molecule weights determine the pharmacokinetics and solubility, which is really important for drug-likeness. The number of rotatable bonds describes the flexibility of 3D molecule conformations which is important when binding to the target protein. QED~\citep{bickerton2012quantifying} is a popular drug-likeness quantized scoring function. LogP stands for partition coefficient which is an important physicochemical property when designing drugs. All these property scores are computed by the RDKit package. 

SAScore~\citep{sas}and RA~\citep{ra} are the synthetic accessibility scores we apply here. SAScore is a rule-based method, in which rare substructures are given a penalty for accessing the synthesizability. It also takes the topological complexity of structures into account. 
RAscore is a scoring function based on neural networks. There is plenty of synthetic route planning AI models developed these years, however, they are always time-consuming for larger datasets. RAscore is a classification model trained upon retrosynthetic models. It predicts the probability of a specific structure having a successful route planned by AI agents.

The validity in Tab.~\ref{sas} stands for the proportion of seed molecules the model is able to generate valid analogues. The maximum depths of tree searching are set to 15 as in the original code repository for SynNet. For the DINGOS (condition) model, we regard the generation process as invalid when the input seed molecules do not match with any template reactions for further modifications. In template-free models, the invalid generations represent that the predictor cannot output valid molecule smiles or one of the reactants is output as the predicted results.

In addition to validity, the distance in Tab.~\ref{sas} is applied to measure the similarity between generated analogues and input seed molecules. The distance is measured by the mean Jaccard distance on the ECFP4 fingerprint of the analogues and the input seed molecules.

We also use the scaffold diversity to measure the generated molecules, which is defined as the information entropy as follows:
\begin{equation}
    H(S) = -\sum_{i=1}^n P(S_i)\log P(S_i)
\end{equation}
where $H(S)$ represents the entropy of a random molecule $S$ with possible scaffold $S_1, S_2, ..., S_n$, and $P(S_i)$ is the probability of a molecule having certain scaffold $S_i$. Intuitively,  a high entropy indicates that the scaffolds are evenly distributed, while a low entropy indicates that there are one or a few dominant scaffolds. 

\textcolor{black}{
We further evaluate the molecule diversity of the generated molecules from different ML models. The diversity is defined as follows:
\begin{equation}
    \operatorname{Div}=\frac{2}{n(n-1)} \sum_{x \neq x^{\prime} \in \mathcal{G}} 1-\operatorname{sim}\left(x, x^{\prime}\right)
\end{equation}
Where $\mathcal{G}$ is defined as the set of all generated molecules and $sim$ is defined as the ECFP4 Tanimoto similarity. 
}

\subsubsection{Metrics for Reaction Retrieval}
\textcolor{black}{\textbf{Enrichment Factor(EF)} is also a widely used metric, which is calculated as}
\begin{equation}
    {\mathrm{EF}}_\alpha=\frac{{\mathrm{NP}}_\alpha}{{\mathrm{NP}}_t\times\alpha}
\end{equation}
\textcolor{black}{Where $\mathrm{NP}_\alpha$ stands for the positive samples in the top $\alpha$ predictions and $\mathrm{NP}_t$ stands for the positive samples in the whole test set. This metric does not rely on the choice of threshold. $\mathrm{EF}_\alpha$ is a measurement of the enrichment of positive samples in the top $\alpha$ predictions.}

\textcolor{black}{\textbf{AUROC}, or Area Under the Receiver Operating Characteristic Curve, is a widely used metric in binary classification. It quantifies the model's ability to distinguish between positive and negative samples by calculating the area under the ROC curve. The AUROC value ranges from 0 to 1, with higher values indicating better classification performance.}

\subsubsection{Case Study}

We conduct molecule docking on  all generated or screened molecules to the 3CL$_{\text{Pro}}$ protein pocket using Glide~\citep{friesner2006extraglide}. At first, the protein structure is processed by Protein Preparation Wizard. Bond orders and missing side chains are fixed using Prime in the preparation workflow. Waters with less than three H-bonds to amino acids are removed. The docking grid of size $10\AA \times 10\AA \times 10\AA$ in which the center is defined by the co-crystal ligand is saved based on the processed protein. On the other hand, the ligands' conformations and possible tautomers and protonation states are generated by the LigPrep module. The ligand conformations are docked into the pocket using Glide~\citep{friesner2006extraglide}. In the meantime, default settings are applied to other parameters. The Glide docking scores are used to rank the results.

\section*{Declarations}

\begin{itemize}
\item Data availability

The USPTO\_MIT and Schneider dataset is downloaded from the Open Reaction Database~(\href{https://docs.open-reaction-database.org/en/latest/}{https://docs.open-reaction-database.org/en/latest/}). We provide our processed training data in python pickle format at \href{https://doi.org/10.5281/zenodo.8075066}{https://doi.org/10.5281/zenodo.8075066}\\
\item Code availability 

The Uni-RXN and Uni-RXN$_\text{gen}$ model in our article are both trained on public datasets. The code to reproduce the results is publicly available at \href{(https://github.com/qiangbo1222/Uni-RXN-official)}{https://github.com/qiangbo1222/Uni-RXN-official}\\

\item Competing interests

The authors declare no competing interests.\\
\item Authors' contributions

B.Q. conceived the initial idea for the projects. B.Q. and Y.D. processed the dataset and trained the model. B.H. provided support on computing resources. B.Q. and Y.Z. performed the experiments using the pre-trained model and the generative model. Y.Z. analyzed the results and B.Q. write the manuscripts. B.Q., S.S., L.Z., and Z.L. contributed to the revising of the manuscripts. The project was supervised by L.Z. and Z.L. All authors took part in discussions.\\

\end{itemize}

\bmhead{Acknowledgments}
This work was financially supported by National Key R\&D Program of China (grant no.2022YFF1203003, grant no.2022YFC2303700), Beijing AI Health Cultivation Project (grant no.Z221100003522022), Peking University Health Science-StoneWise Technology Joint Laboratory Project (grant no.L202107), the Open Fund of State Key Laboratory of Pharmaceutical Biotechnology, Nanjing University, China (grant No. KF-202304).

\bibliography{sn-bibliography}

\newpage
\begin{appendices}

\section{Related Works Discussion}
 In this section, we discussed two related branches of related work, chemical reaction pretraining and reaction-based molecule generation. 

There has been plenty of works that address the task of representing the chemical reaction. \textcolor{black}{Studies have addressed the task of representing chemical reactions, focusing on the differences between traditional molecule fingerprints like Morgan Fingerprint and Atom Pair Fingerprint, which capture subgraph changes in reactions~\citep{schneider2015development}.} Later, based on the hypothesis that chemical reaction is the rearrangement of different functional groups, rule-based representations~\citep{probst2022reaction} are developed, and they can be applied to the classification of reactions. To further improve the performance on some harder downstream tasks, deep learning based representations are proposed. At first, a large language model~\citep{irwin2022chemformer} that pretrained on individual molecule structures was introduced to enhance the performance of reaction prediction. Rxnfp~\citep{schwaller2021mapping} is the first pretraining framework on reactions in which a BERT-like~\citep{bert} mask language learning method is applied directly to build models on chemical reaction smiles sequence from a commercial reaction database. With the help of large-scale hand-curated data, the model achieves impressive outcomes. 
\textcolor{black}{However, molecule graph provides more detailed and explicit representations than smiles, such as connectivity and stereochemistry.}
Therefore, a few research~\citep{wen2022improving, wang2021chemical} adapted graph neural networks to chemical reactions using contrastive learning. \textcolor{black}{In our approach, we utilize multimodal representations for reactants and reagents and train both graph-based and text-based transformers to derive comprehensive representations of chemical reactions.}

\textcolor{black}{Besides representation learning}, the molecule generative model based on chemical reactions is also an important application of deep learning in the chemical reaction space. Generative models aim to sample similar but different data points from the original data distribution, which can be applied to develop large-scale drug-like molecule vocabulary. In order to make the generated structure accessible, the chemical reactions-based method is designed to generate molecules with desired properties~\citep{korovina2020chembo, bradshaw2019model, bradshaw2020barking}. For example, ChemBO~\citep{korovina2020chembo} applied Bayesian optimization in the latent space to find the desirable molecules. However, in most cases which property is directly related to drug likeness is unclear. Hence, models~\citep{gao2021amortized, button2019automated, pmlr-v162-noh22a} have been proposed to generate analogues based on the input target structure for ligand-based drug design. For example,  C-RSVAE~\citep{pmlr-v162-noh22a} is developed to generate synthetic routes for molecule generation using reaction templates and fragment building blocks, which generate large numbers of molecules with desirable properties. 
\textcolor{black}{However, template reactions introduce undesirable factors. Extensive manual extraction are required to derive high-quality reactions~\citep{synthis} In addition, template reactions limit the model's capacity to generate directly based on the seed structure, which will be discussed in Sec.~\ref{gen_sec} }
To solve the above problems, we propose our generative framework that is able to directly generate novel molecules similar to the seed molecule in a template-free fashion.

\section{More Implementation Details}\label{secA1}
\subsection{Reaction Network and Reaction Path Dataset}
\begin{figure}[h]
\centering
\includegraphics[width=1\textwidth]{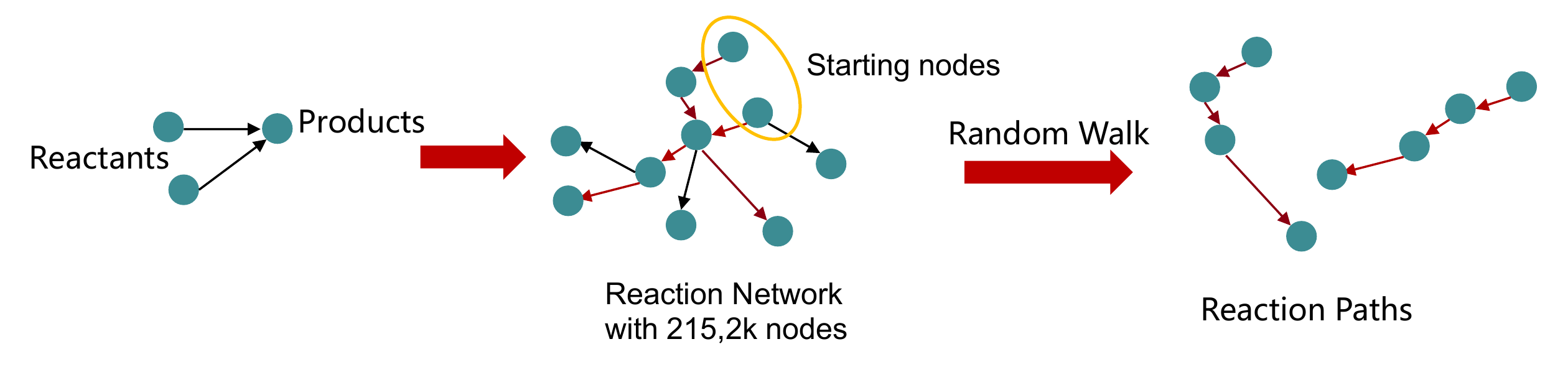}
\caption{An illustration of the workflow to extract reaction paths dataset to train our conditional generation model} \label{fig:net} 
\end{figure}
\bigskip

Inspired by previous work on the reaction network~\citep{jacob2018statistics}, we first construct a chemical reaction network that consists of 2,152k nodes. Each pair of connected nodes corresponds to a chemical reaction from the USPTO STEREO dataset. The original data provides atom mapping information, so we extract the main reactants that possess the most atom mapping with the product from all reactants as the starting points of the edges. The endpoints of edges are the products of reactions. The sub-reactants and reagents are used as representations of the edges between the main reactants and products. In other words, every path in the graph represents a series of reactions that travels from one structure to another structure. All the paths derived from our reaction network are guaranteed to be valid and can be produced in wet labs. 

Then, we sample paths in the graph to build our training / valid / test datasets. We select all the nodes that have a QED score higher than 0.5 to be our starting points to filter unstable reactive structures. From these nodes, we apply random walk sampling with depths from 1 to 3 in order to obtain diverse paths. Finally, we derived 49039 valid chemical paths. The whole data processing procedure is shown in Fig~.\ref{fig:net}. The orders of the sampled paths are reversed to train the model. All the chemical paths are split as pairs of the former path and next step response~(sub-reactants/ regents), so these pairs can be used for teacher force training. The inputs are the former sequences in the chemical paths, while our target is to imitate the real reactants for the coming chemical reaction editing.

\subsection{Detailed Implementation of Generative Model}

\textcolor{black}{Our generative model is designed to approximate the representations of target responses and predict the termination of reaction paths. To achieve this, we employ four loss terms during training using gradient descent. The first loss term measures the distance between the representations of the responses generated by the Set Generator and the target representations. We utilize the inner product between representations as a distance metric and train on the in-batch contrastive objective. The second loss term involves binary cross-entropy loss, which compares the predicted selected index of $\boldsymbol{X}$ with the ground truth number of responses. These two loss term is denoted as the reconstruct-loss in our Algorithm~\ref{alg:train}. In addition, we incorporate a loss term to guide the model in learning the termination of reaction editing paths. An MLP is used to predict whether the current set of responses is the final one. Binary cross-entropy is employed to train this termination network. The final loss term is the KL divergence between $z_p$ and $z_r$, as illustrated in Fig.~\ref{gen_all}\textbf{b}.}

\textcolor{black}{During the inference phase, our model generates complete reaction paths using the following steps, as in Algorithm~\ref{alg:sample}. Firstly, the input molecule structure is encoded into a representation using our pre-trained encoder and processed by the LSTM network. The prior network then outputs $z_p$, which is used for generating the response set. Conditioned on $z_p$, $n$ representations are generated. We employ the fiass package~\citep{fiass} developed by Facebook to search within our precomputed reactant/reagent representation library, which consists of millions of entries. The forward product predictor takes the generated reactants/reagents as input and predicts the intermediate structure. If the network predicts that it is time to terminate the generation, the final product is output. Notably, our model allows users to define their own termination criteria, such as keeping the molecule weight below a specified threshold. If our network does not predict termination, the intermediate structure is appended to the reaction path for the next-step generation. }

\renewcommand{\algorithmicrequire}{\textbf{Input:}}
\renewcommand{\algorithmicensure}{\textbf{Output:}}
\label{alg:train}
\begin{algorithm}
    \caption{Training Algorithm for Our Conditional Variational Auto-Encoder}
    \begin{algorithmic}[1]
    \Require{Reaction Paths Queue $P$, Pretrained Encoder $\phi_{r}$}
    \Ensure{Path LSTM Network $\phi_{lstm}$, Recognition Network $\phi_{\text{rec}}$, Prior Network $\phi_{\text{pri}}$, Generator $\operatorname{G}$}

    \For {i in epochs}
        \For {path in $P$}
        \State Sample random number l $<$ length of path
        \State input path = path[ :l], target response = path[l]
        \State former context = $\phi_{lstm}$($\phi_{r}$(input path))
        \State target representation = $\phi_{r}$(target response)
        \State $z_{\text{pri}}$ = $\phi_{\text{pri}}$(former context)
        \State $z_{\text{rec}}$ = $\phi_{\text{rec}}$(former context, target representation)
        \State predicted representation = $\operatorname{G}$($z_{\text{rec}}$, former context)
        \State Loss = Reconstruct-loss(predicted representation, target representation) + KL-divergence($z_{\text{pri}}$, $z_{\text{rec}}$)
        \State Update $\phi_{lstm}$, $\phi_{\text{rec}}$, $\phi_{\text{pri}}$, $\operatorname{G}$ $\leftarrow$ Optimize Loss
        \EndFor
    \EndFor
    \end{algorithmic}
\end{algorithm}

\label{alg:sample}
\begin{algorithm}
    \caption{Sampling Algorithm for Our Conditional Variational Auto-Encoder}
    \begin{algorithmic}[1]
    \Require{Seed Structure $m$, Pretrained Encoder $\phi_{r}$, Path LSTM Network $\phi_{lstm}$, Prior Network $\phi_{\text{pri}}$, Generator $\operatorname{G}$, Reaction Predictor $\operatorname{RP}$, Sub-reactants/ Reagents Library $L$, max length, termination criterion}
    \Ensure{Analogue $a$}
    \State path = \{$m$\}
    \State Lib = \{$\phi_{r}$($l$) $\mid l \in L$\}
    \While{path length $<$ max length}
        \State former context = $\phi_{lstm}$($\phi_{r}$(path))
        \State $z_{\text{pri}}$ = $\phi_{\text{pri}}$(former context)
        \State generated representation = $\operatorname{G}$($z_{\text{pri}}$, former context)
        \State generated responses $\leftarrow$ Search within Lib
        \State path = \{path, $\operatorname{RP}$(path[-1], generated responses)\}
        \If{termination criterion satisified}
            \State \textbf{Break}
        \EndIf
    \EndWhile
    \State a = path[-1]
    \end{algorithmic}
\end{algorithm}

\subsection{Hyperparameters}
 The hyperparameters of our model are chosen through a combination of manual tuning and automated searching. The hyperparameters for the model architecture are set manually. For both the graph-based transformer and text-based transformer that we employ in the shared encoders, we apply a 4-head network with a width of 512 dimensions. The transformer model we apply in the reactive center prediction task is reduced to a 2-layer model with the same number of attention heads and width dimensions. In all graph-based transformer models, we set the multi-hop edge model with a maximum distance cut-off of 6. The hidden sizes of all projection heads are 512. In the invariant generator, we set the channels of reference set to 512 and the dimension of the latent variable to 1024. The $z$ dimension for the recognition network and prior network is set to 512. The hidden dimension of MLPs we employed in the invariant generator is also set to 512. The threshold $\delta$ for selecting data points is set to 0.5.

The batch size and learning rate are tuned by the automatic optimum finder from Pytorch Lightning. A learning rate scheduler that shrinks the learning rate to half of its value at the end of every 20 epochs is adopted too. We apply an early stopping strategy based on the loss curve of loss on the validation set.

\subsection{Implementations of baselines}
 We evaluate our model compared to the baseline models. We include 3 methods that focused on pretraining or representation learning.

\textbf{Rxnrep}~Inspire by contrastive learning in computer visions~\citep{he2020momentum}, a graph neural network model is pre-trained using a self-supervised training objective~\citep{wen2022improving}. The embedding similarity between positive samples is maximized, while the ones between negative samples are minimized. The positive samples are drawn by masking atoms/ bonds/ subgraphs.

\textbf{MolR}~Instead of contrastive learning on positive/ negative samples, MolR~\citep{wang2021chemical} applied training objective that maximized the embeddings between the input of the reaction (reactants/ reagents) and the output of the reaction (products). Multiple graph neural network backbones were tested in the original article. We chose the graph attention networks(GAT) backbone since it outperforms other backbones in multiple experiments.

\textbf{DRFP}~This fingerprint is a model-free representation~\citep{probst2022reaction}, which used a combination of two kinds of chemical fingerprints ECFP and MHFP. The circular substructures and the subsequent hashing are used to represent each molecule structure within the chemical reactions.

We include the following generative models to evaluate our generative model's performance.

\textbf{SynNet}~A tree generation model is proposed by previous work~\citep{gao2021amortized}. The model is trained on artificial pathways generated from purchasable compounds and templates. We apply this model to structural analogs. However, the model is not capable to generate analogs for all drug structures. The tree-searching algorithm cannot construct any similar valid structure within the max searching depth.

\textbf{DINGOS}~The code of DINGOS model is derived from the original article~\citep{button2019automated}. Two versions of the DINGOS model are proposed. The DINGOS (\textit{De novo}) model chooses the starting material based on a similarity function to search for the most suitable purchasable molecule. The DINGOS (condition) takes our input drug structure as the starting material and generates one step of reaction modification based on the neural network predicted template reaction and building blocks. The product of the template reaction is collected as the structure analog.

\textcolor{black}{\textbf{Lib-INVENT}~The model of this library enumeration tool is derived from a previous work~\citep{libinvent}. Instead of based on chemical reactions, Lib-INVENT decomposed the molecule structures from the training set using BRICS rules and approximate the distribution of the masked fragments.} \textcolor{black}{Following the Lib-INVENT paper, we used BRICS rules to cut the drug structures and keep the largest fragment as the input scaffolds for the model. To conduct a fair comparison, we do not apply the reinforcement learning techniques and only use the prior model which is an unbiased molecule generative model.}


\section{Additional Experiments}\label{secA2}
\subsection{TPL 1k Classification Task}
 Following previous work~\citep{schwaller2021mapping}, we test our model on the TPL 1k reaction classification dataset. This is a rule-based classification system and the benchmark dataset is unbalanced. In addition to accuracy, we also evaluate the confusion entropy of the CEN metric which provides a fair comparison of multi-class classification tasks. When we denote Matrix as the confusion matrix, the CEN can be derived as follows:

\begin{align}
P_{i, j}^{j} & =\frac{\operatorname{Matrix}(i, j)}{\sum_{k=1}^{|C|}(\text { Matrix }(j, k)+\operatorname{Matrix}(k, j))} \nonumber\\
P_{i, j}^{i} & =\frac{\operatorname{Matrix}(i, j)}{\sum_{k=1}^{|C|}(\text { Matrix }(i, k)+\operatorname{Matrix}(k, i))} \nonumber\\
\mathrm{CEN}_{j} & =-\sum_{k=1, k \neq j}^{|C|}\left(P_{j, k}^{j} \log _{2(|C|-1)}\left(P_{j, k}^{j}\right)+P_{k, j}^{j} \log _{2(|C|-1)}\left(P_{k, j}^{j}\right)\right) \nonumber\\
P_{j} & =\frac{\sum_{k=1}^{|C|}(\operatorname{Matrix}(j, k)+\operatorname{Matrix}(k, j))}{2 \sum_{k, l=1}^{|C|} \operatorname{Matrix}(k, l)} \nonumber\\
\mathrm{CEN} & =\sum_{j=1}^{|C|} P_{j} \mathrm{CEN}_{j}\nonumber
\end{align}

The results are shown in Table.~\ref{tpl}. Our model shows comparable accuracy with other methods and outperforms other baseline methods on the CEN metrics. These results illustrate that our model not only performs well on small datasets but also is able to provide comparable performance on large datasets. rxnfp\citep{schwaller2021mapping} outperforms our model by 6.2\%, when pretraining on a commercial dataset that possesses 2 million hand-curated data instead of an openly accessible database that possesses 500k noisy data. DRFP outperforms our model by using a rule-based fingerprint, which takes short-cut when benchmarking on a rule-based classification task.
\begin{table}[h]
\begin{center}
\begin{minipage}{0.5\textwidth}
\caption{The experimental results on the TPL 1k dataset}\label{tpl}

\begin{tabular*}{\textwidth}{@{\extracolsep{\fill}}ccc@{\extracolsep{\fill}}}
\toprule%
Methods & Accuracy & CEN\\
 \midrule
AP3 256 (MLP) & 0.809 & 0.101 \\
AP3 256 (5-NN) & 0.295 & 0.242 \\
DRFP (5-NN) & 0.917 & 0.041 \\
DRFP (MLP) & 0.977 & 0.011\\
rxnfp & 0.989 & 0.006 \\
hypergraph & 0.928 & --\footnotemark[1] \\
\midrule
Uni-RXN & 0.927 & 0.003 \\

\botrule
\end{tabular*}
\footnotetext[1]{The CEN results are missing because the original paper does not provide the code to reproduce the predictions.}
\end{minipage}
\end{center}
\end{table}

\subsection{Evaluation on Different Negative Data Sampling Mechanisms}\label{si_neg}
\textcolor{black}{
In addition to our proposed approach discussed in Section~\ref{our_neg_method}, various other methods have been proposed to generate unoptimized chemical reactions. One such method was introduced in a previous work~\citep{filter}, which presented three different mechanisms based on template enumeration. Among these mechanisms, we specifically opted for the 'random' mechanism as discussed in the original paper, as it not only generates perturbed products but also perturbed sub-reactants.}

\begin{figure}[t]
\centering
\includegraphics[width=1.0\textwidth]{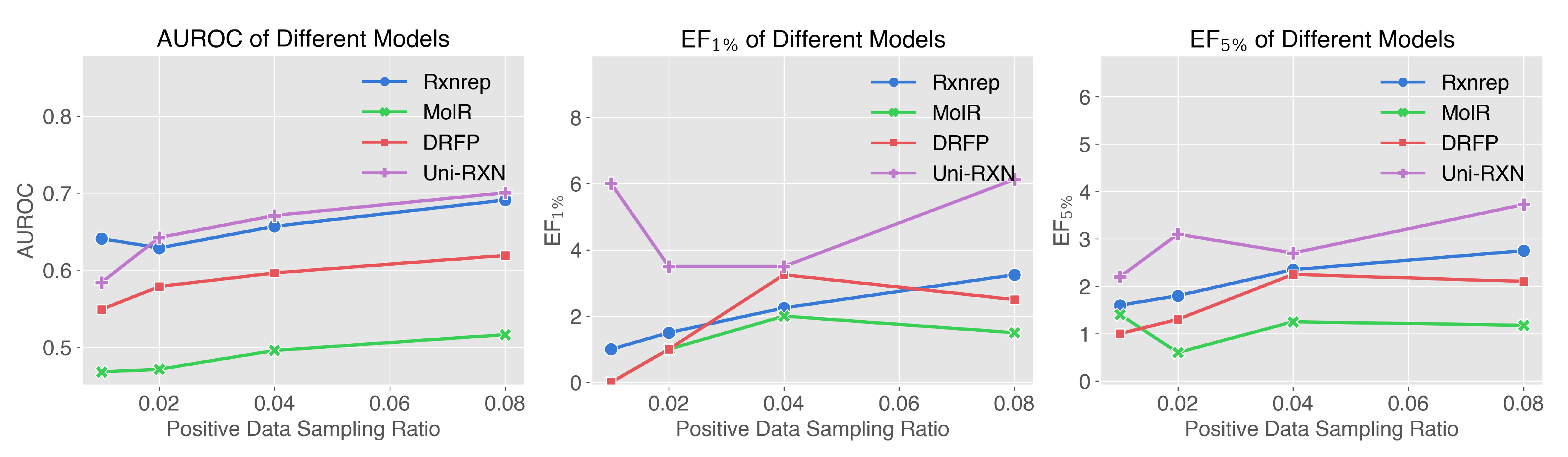}
\caption{\textcolor{black}{The line charts illustrate the AUROC, EF$_1$$_{\%}$ and EF$_1$$_{\%}$ of different representations in the chemical reaction retrieval task.}} \label{abla_aiz}
\end{figure}

\textcolor{black}{To further investigate the performance of our model in enriching positive data among the top-ranked reactions, we sampled different ratios of positive data and conducted experiments. The results are depicted in Fig~\ref{abla_aiz}. It is evident that compared to our method, the template-based mechanism employed in this study makes it considerably easier for the model to differentiate between optimized and unoptimized reactions. Despite this, Uni-RXN continues to outperform other reaction representations even in this different setting.}

\bigskip

\subsection{Molecule Docking Protocol Validation}

\begin{figure}[t]
\centering
\includegraphics[width=1.0\textwidth]{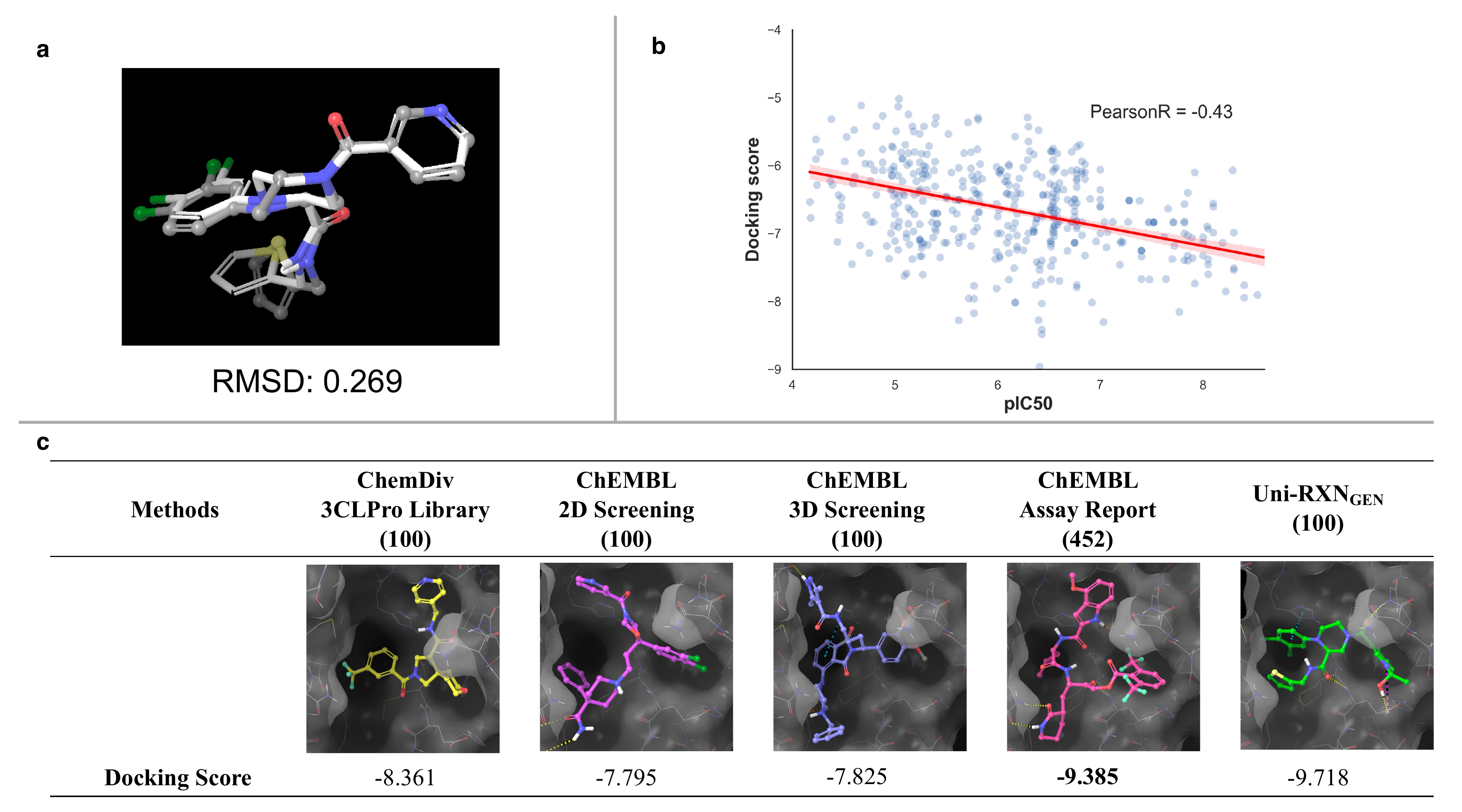}
\caption{\textcolor{black}{\textbf{(a)} The conformation of experimental co-crystal conformation and the redocked conformation \textbf{(b)} The correlation between the experimental binding affinity with the docking score. Most assay data points are close to the red line indicating that the structures with higher docking score absolute value are much more likely to bind to the protein target pocket.  \textbf{(c)} The docking pose of the top scoring screened COVID-19 3CLPro inhibitors.}} \label{fig:case_study} 
\end{figure}

\textcolor{black}{At first, we redocked the co-crystal molecule from a previous study~\citep{gao2022discovery}, and we also performed docking for all molecules with reported IC50 values towards the target protein~(PDB id: \href{8ACL}{http://doi.org/10.2210/pdb8ACL/pdb}) where the data are collected from ChEMBL~\citep{bento2014chembl}. The redocked conformation is depicted in Fig~\ref{fig:case_study}(a) and the correlation between pIC50 and docking scores is presented in Fig~\ref{fig:case_study}(b). The RMSD result demonstrates that molecule docking can provide high-quality binding conformations, while the correlation validates the suitability of docking scores in representing potential binding affinity.}

\textcolor{black}{Besides deep learning models we evaluated in the main text, we also compared our model with other virtual screening methods. The ChemDiv 3CL$_{\text{Pro}}$ Library is a commercial dataset that is designed particularly for this protein target, we choose the 100 molecules according to the ECFP4 similarity scores to the reference molecule. We further use two ligand-based methods to carry out virtual screening on the full ChEMBL dataset. 2D screening is also based on the ECFP4 Jaccard distance, the top 100 most topological similar structures are selected. The 3D screening method is based on the 3D pharmacophore, in which the Develop Pharmacophore Model module was used to generate pharmacophore hypotheses based on the reference structure, and the Phase Ligand Screening module was used to perform pharmacophore screening. Both modules are provided by Schrodinger 2018 suite. As a result, 15249 molecules hit all pharmacophore features, and the top 100 molecules were selected for virtual screening.}

\textcolor{black}{The results have shown that compared to the top-scoring molecules searched by these virtual screening approaches, our model is still able to find the structure with the highest potential binding affinity.}

\subsection{\textit{De novo} Generation}
\textcolor{black}{Furthermore, our model exhibits versatility beyond its primary focus on structure conditional molecule generation, as it can also be effectively applied to \textit{de novo} molecule generation. In the case of \textit{de novo} generation, our model takes a different approach by randomly selecting small fragments from the ZINC fragment-like library~\citep{irwin2020zinc20} as initial seeds. An additional MLP was trained to predict the molecule weight of the final output. With this auxiliary module, our model terminates the generation once the intermediate products have reached the predicted molecule weight. We evaluate the quality of the generated molecules using the well-established MOSES metric\citep{moses} and compare the results with two other reaction-based generative models trained on the USPTO dataset. To ensure a fair comparison, we employ the same reaction predictor~\citep{schwaller2019molecular} used by these baseline models. The results, presented in Table~\ref{tab_denovo}, demonstrate that our model outperforms the baseline models in terms of Novelty and scaffold int-Div metrics. This indicates that our model has the ability to generate diverse scaffolds and novel structures, solidifying its strength in \textit{De novo} molecule generation.}

\begin{table}[h]
\begin{center}
\begin{minipage}{0.92\textwidth}
\caption{\textcolor{black}{MOSES metrics for \textit{De novo} generation}}\label{tab_denovo}

\begin{tabular*}{\textwidth}{@{\extracolsep{\fill}}cccccc@{\extracolsep{\fill}}}
\toprule%
Model & Validity & Unique & Novelty & int-Div & Scaffold int-Div \\
 \midrule
Molecule Chef~\citep{bradshaw2019model}& 98.9\% & 99.0\% & 90.0\% & 0.888 & 0.850 \\
DAG-Gen~\citep{bradshaw2020barking} & 100\% & 99.0\% & 88.4\% & 0.880 & 0.843 \\
\midrule
Uni-RXN$_\text{Gen}$ & 100\% & 97.6\% & 100.0\% & 0.875 & 0.902\\

\botrule
\end{tabular*}
\end{minipage}
\end{center}
\end{table}

\subsection{Ablation Study}
\paragraph{Pretrained Model}

\textcolor{black}{Our pretrained model has undergone rigorous training on an extensive dataset encompassing a diverse array of chemical reactions. To ascertain its ability to generalize and maintain robustness when faced with previously unseen classes of chemical reactions, we conducted a meticulous ablation study. In this study, we performed various modifications, such as removing specific pretraining tasks or randomly subsampling a smaller pretraining dataset, and subsequently evaluated the model's performance on the same Schneider test set. The comprehensive results of these ablation experiments are meticulously documented in Tab.~\ref{abla-pre}, providing valuable insights into the model's adaptability and resilience in diverse scenarios.}

\textcolor{black}{Furthermore, we sought to gauge the model's proficiency in a more specific context by selectively excluding all reactions belonging to the heteroatom alkylation and arylation superclass from the pretraining dataset. Subsequently, we exclusively utilized reactions from this particular superclass for testing purposes. By doing so, we were able to assess the model's accuracy in predicting the correct secondary classification for these restricted reactions. The corresponding accuracy values, captured in Tab.~\ref{abla-pre}, shed light on the model's performance and reliability when confronted with a distinct chemical reaction category.}

\textcolor{black}{By conducting these meticulous evaluations and ablations, we aim to provide comprehensive insights into the generalization capabilities and robustness of our pretrained model. These findings not only contribute to a deeper understanding of the model's performance characteristics but also highlight its potential applicability in handling novel chemical reaction classes encountered in real-world scenarios.}

\begin{table}[h]
\begin{center}
\begin{minipage}{\textwidth}
\caption{\textcolor{black}{Ablation study results of the pretraining tasks and size of dataset}}\label{abla-pre}

\begin{tabular*}{\textwidth}{@{\extracolsep{\fill}}cccccc@{\extracolsep{\fill}}}
\toprule%
MS Loss& RP Loss& Center Loss & dataset size & Ex Overlap & Accuracy\\
\midrule
$\checkmark$ & $\checkmark$ & $\checkmark$&100\% & - & $0.600\pm0.0293$\\
\midrule
$\checkmark$ & $\checkmark$ & $\checkmark$&100\% & $\checkmark$ & $0.587\pm0.0229$\\
\midrule
$\checkmark$ & $\checkmark$ & $\checkmark$&10\% & $\checkmark$ & $0.468\pm0.0242$\\
\midrule
$\checkmark$ & $\checkmark$ & - &100\% & $\checkmark$ & $0.540\pm0.0145$\\
\midrule
$\checkmark$ & - & - &100\% & $\checkmark$ & $0.485\pm0.0243$\\

\botrule
\end{tabular*}
\footnotetext{\textcolor{black}{MS stands for 'Main Reactant-(sub-reactant reagents) Pairing' and RP stands for '(Reactants, reagents)-product Pairing'. Ex Overlap stands for the pretraining dataset in which all the reactions whose templates can be found in our test set are removed. The accuracy results are evaluated when only 4 reactions are kept per reaction class for both the finetuning dataset and the test dataset.}}
\end{minipage}
\end{center}
\end{table}

\begin{table}[h]
\begin{center}
\begin{minipage}{\textwidth}
\caption{\textcolor{black}{The accuracy of the heteroatom alkylation and arylation superclass reaction classification}}\label{abla_heter}
\begin{tabular*}{\textwidth}{@{\extracolsep{\fill}}ccccc@{\extracolsep{\fill}}}
\toprule%
 \makecell{Reaction Number \\per Class} & Rxnrep  & MolR  & DRFP  & Uni-RXN \\
\midrule
4 & $0.150\pm0.0214$ & $0.243\pm0.0513$ & $0.193\pm0.0497$ & $\mathbf{0.450\pm0.0653}$ \\
8 & $0.282\pm0.0578$ & $0.314\pm0.0497$ & $0.258\pm0.0407$ & $\mathbf{0.629\pm0.0230}$ \\
16 & $0.352\pm0.0148$ & $0.390\pm0.0245$ & $0.337\pm0.0252$ & $\mathbf{0.710\pm0.0411}$ \\
Full & $0.626\pm0.0015$ & $0.829\pm0.0009$ & $0.675\pm0.0011$ & $\mathbf{0.901\pm0.0013}$ \\
\botrule
\end{tabular*}
\footnotetext{The accuracy is computed multiple times on different random samples. The Standard Deviation(std) of the accuracy is listed after $\pm$. The higher accuracy indicates better performance. \textcolor{black}{Trained graph neural network encoder from Rxnrep~\citep{wen2022improving}, MolR~\citep{wen2022improving} is applied to compute the baseline model representation. The package that computes DRFP~\citep{probst2022reaction} representation is downloaded directly from rxn4chemistry}}
\end{minipage}
\end{center}
\end{table}

\paragraph{Generative Model}
 In order to verify the function of our pre-trained encoder in the generative model, we carry out an ablation study to prove that these pretraining tasks aid our generative model in generating sub-reactants and reagents accurately. We removed the shared encoder in different stages and replaced it with encoders that are trained from scratch or other fingerprint encoding. Because different representation is used to encode the target structure, it is not suitable to measure the quality using representation distances. Therefore, we randomly select a batch of other reactants/ reagents with different representations encoded and rank the ground truth target representation among this batch of representations. A Low ranking number indicates that our model generates representations that are more likely to retrieve the plausible reactants/ reagents from the library. The results are listed in Table.~\ref{abla}.

\begin{table}[h]
\begin{center}
\begin{minipage}{\textwidth}
\caption{Ablation study results of the model with different encodings of target and input structures}\label{abla}
\begin{tabular*}{\textwidth}{@{\extracolsep{\fill}}cccc@{\extracolsep{\fill}}}
\toprule%
Uni-RXN Input & Uni-RXN Target & Pretrained Encoder & Ranking scores\\
\midrule
-&- &- & 0.3474\\
\midrule
$\checkmark$&-&-&0.2451\\
\midrule
-&$\checkmark$&$\checkmark$&0.2304\\
\midrule
$\checkmark$&$\checkmark$&-&0.3486 \\
\midrule
$\checkmark$&$\checkmark$&$\checkmark$&0.1243\\

\botrule
\end{tabular*}
\footnotetext{The lower ranking scores indicate better performance. We used ECFP4 as the encodings for molecule structures if we remove the pre-trained encoders.}
\end{minipage}
\end{center}
\end{table}

\bigskip

The results illustrate that without the pre-trained encoder, the generative model is not able to approximate the data distribution of the corresponding sub-reactants/ reagents. When we use ECFP4 as the fingerprints to encode our molecule structure, our model's ranking scores increase which means the fingerprint-based model fail to differentiate molecules that play different roles in chemical reactions. However, this encoding strategy is applied in a few related works~\citep{gao2021amortized, button2019automated}. In conclusion, the ablation study shows that our Uni-RXN embedding is a good representation of chemical entities in the reactions. The trained encoders help our model generate responses that maintain the consistency of the input structures.

\subsection{Yield Regression Experiment}\label{appdx:yield}
\begin{figure}[t]
\centering
\includegraphics[width=1\textwidth]{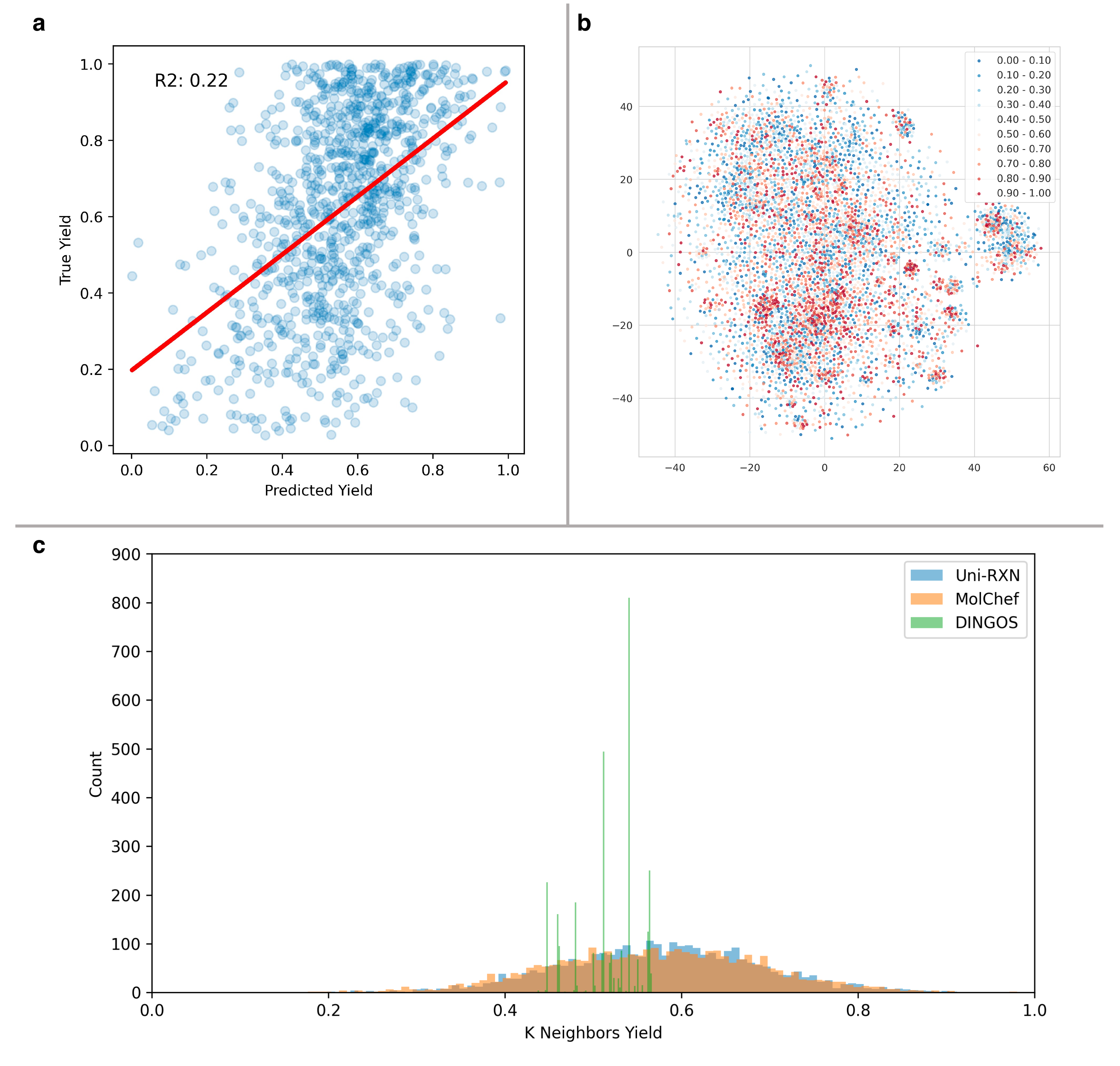}
\caption{\textbf{(a)}~The relation between our model predicted K-Neighbor yield and ground truth yield. \textbf{(b)}~The T-SNE visualization of the Uni-RXN chemical reaction space colored with yield data. \textbf{(c)}~The distribution of the chemical reactions' predicted yields proposed by different methods.} \label{fig:yield} 
\end{figure}
\bigskip

 First, we design a weighted K-Neighbor method to predict the yield of the chemical reactions. It is hard to predict the chemical reaction yields for the open patent dataset because the dataset is noisy and has an unsmooth yield landscape~\citep{schwaller2021prediction}. We use a weighted K-neighbor method to predict the yield. Specifically, we search for the K nearest neighbors in the chemical reaction space encoded by Uni-RXN encoders. If we denote the K nearest distances as $D$ and the yield of these K reactions as $L$. The predicted yield $y$ can be derived as:
\begin{equation}
    y = \frac{\sum_{i\in 1\cdots K}(1/D_{i} * L_{i})}{\sum_{i\in 1\cdots K}(1/D_{i})}
\end{equation}

When we exclude 1000 data from the USPTO dataset to test our method's ability. The result is illustrated in Fig.~\ref{fig:yield}(a). In the only previous work~\citep{schwaller2021prediction} that tested its performance to predict yield on the USPTO dataset, they achieved results of $R_2$ less than 0.2 on the gram and sub-gram dataset. Our model clearly outperforms their methods by a large margin and is suitable for predicting the approximation of the reaction yield. The visualization results in Fig.~\ref{fig:yield}(b) demonstrate that though the landscape is still unsmooth, the well-optimized reactions with high yields ranging from 90\% to 100\% are clearly well clustered in our chemical reaction space.

Based on this K-neighbor method, we predict the yield for the reactions generated by the template-based method and the template-free method. The results are plotted in Fig.~\ref{fig:yield}(c). We can see that the template reactions with the green color in Fig.~\ref{fig:yield}(c) are not well-optimized. However, the template-free methods (Uni-RXN, MolChef) generate more high-yield reactions. The average predicted yield for the reactions generated by our model is 57.68\%. For MolChef model~\citep{bradshaw2019model}, the predicted yield is 57.02\%. Our method outperforms both another template-free model and the template-based method.




\end{appendices}


\end{document}